\documentclass{article}

\PassOptionsToPackage{numbers}{natbib}
\usepackage[preprint]{neurips_2025}

\usepackage[utf8]{inputenc} % allow utf-8 input
\usepackage[T1]{fontenc}    % use 8-bit T1 fonts
\usepackage{hyperref}       % hyperlinks
\usepackage{url}            % simple URL typesetting
\usepackage{booktabs}       % professional-quality tables
\usepackage{amsfonts}       % blackboard math symbols
\usepackage{amssymb}        % for symbols like \leftrightarrow
\usepackage{multirow}       % for multi-row cells in tables
\usepackage{nicefrac}       % compact symbols for 1/2, etc.
\usepackage{microtype}      % microtypography
\usepackage{xcolor}         % colors
\usepackage{algorithmic}
\usepackage{algorithm}

\usepackage{subfigure}
\usepackage{caption}
\usepackage{booktabs} % for professional tables
\usepackage{comment}

\usepackage{amsmath}
\usepackage{amssymb}
\usepackage{mathtools}
\usepackage{amsthm}

\newtheorem{theorem}{Theorem}[section]

\theoremstyle{definition}
\newtheorem{definition}[theorem]{Definition}

\theoremstyle{remark}
\newtheorem{remark}[theorem]{Remark}

\DeclareMathOperator*{\argmin}{argmin}
\newcommand{\ant}{\ensuremath{\texttt{ANTIDOTE}}}

\title{Learning to Forget with Information Divergence Reweighted Objectives for Noisy Labels}

\author{%
  Jeremiah~Birrell\\
  Department of Mathematics\\
  Texas State University\\
  San Marcos, TX \\
  \texttt{jbirrell@txstate.edu} \\
  \And
  Reza Ebrahimi\\
  School of Information Systems and Management \\
  University of South Florida \\
  Tampa, FL \\
  \texttt{ebrahimim@usf.edu} \\
}

\begin{document}

\maketitle

\begin{abstract}
  We introduce \ant, a new class of objectives for learning under noisy labels which are defined in terms of a relaxation over an information-divergence neighborhood. Using convex duality, we provide  a reformulation as an adversarial training method that has similar computational cost to training with standard cross-entropy loss. We show that our approach adaptively reduces the influence of the samples with noisy labels during learning, exhibiting a behavior that is analogous to forgetting those samples. \ant~ is effective in practical environments where label noise is inherent in the training data or where an adversary can alter the training labels. Extensive empirical evaluations on different levels of symmetric, asymmetric, human annotation, and real-world label noise show that \ant~outperforms leading comparable losses in the field and enjoys a time complexity that is very close to that of the standard cross entropy loss.\end{abstract}

\section{Introduction} 
Training effective machine learning models is challenging in many practical and critical environments where label noise is inherent in the training data or in cases of poisoning attacks in which an adversary may alter the labels of the training set. Modified losses can help the model be less affected by label noise during training \citep{ ma2020normalized_normalized, zhou2021asymmetric_loss, zhu2023label_LDR_KL_ICML, asymmetric_TPAMI_2023, wangepsilon_2024_neurips} and  estimators of the   noise transition matrix can be used for resampling or reweighting \cite{natarajan2013learning,liu2015classification,patrini2017making,xia2019anchor,baedirichlet}. Some studies make the assumption that when the label noise is not dominant, the noise can be ignored in the early training iterations. This view can lead to suboptimal solutions due to the risk of noise memorization in the beginning of the training, also known as the warm-up obstacle \cite{zheltonozhskii2022contrasttodivide_c2d, liu2020early_ELR}. Consequently, many methods rely on additional expensive label correction techniques (e.g., pseudo label prediction, Mixup label interpolation) or semisupervised training to rectify the effect of noise during later training iterations \cite{zhang2018mixup, Li2020DivideMix, han2020robust_NeurIPS_COTeaching, liu2020early_ELR, liu2022robust_SOP, zheltonozhskii2022contrasttodivide_c2d, chang2023csot_labelcorrection_optimaltransport,  sohn2020fixmatch_pseudolabel, zhou2024l2b, zhang2024badlabel_TPAMI}.

Our study offers a novel and principled approach to the long-standing challenge of label noise that does not rely on clean data or any label correction. Drawing on information theory and adversarial training, we first formulate the problem as a relaxation of the expected loss over an information-divergence neighborhood. Under appropriate assumptions, we prove that the true labels are the unique solution to this relaxed problem, even in the presence of label noise.   Then we propose a more computationally efficient reformulation of the objective, which is framed in terms of a class of new loss functions  that are trained in an adversarial manner. As the training proceeds, the result of our method can be viewed as determining and then forgetting the  samples with  noisy labels. We refer to the proposed  new method as {\bf A}dversarial {\bf N}eural-Network {\bf T}raining with {\bf I}nformation {\bf D}ivergence-Reweighted {\bf O}bjec{\bf t}iv{\bf e} (\ant).  These new objectives can be easily integrated into any supervised deep learning model with little added computational cost.

\subsection{Learning with Noisy Labels via Relaxation-Optimization}
Fitting a classifier to a set of labeled data using standard empirical risk minimization is equivalent to solving the  optimization problem
\begin{align}\label{eq:stoch_opt}
\inf_\theta E_{P_n}[\mathcal{L}_\theta]\,,
\end{align}
where $\mathcal{L}_\theta$ is the loss (e.g., cross-entropy between predicted label probabilities and the true label), depending on the model parameters $\theta\in\Theta$,  $P_n=\frac{1}{n}\sum_{i=1}^n \delta_{(x_i,y_i)}$ is the empirical distribution of the training samples $x_i$ with labels $y_i$ (with $\delta_{(x_i,y_i)}$ denoting the delta distribution supported on the point $(x_i,y_i)$), and $E_{P_n}$ is the expectation with respect to $P_n$.    In practice, the training set will often contain some amount of bad data (e.g., mislabeled samples), either intentionally due to the efforts of an attacker or simply due to unintentional human error.   In this work, we propose a relaxation-optimization strategy for identifying the mislabeled samples and reducing their effect on the trained model, i.e., we allow the method to (partially or completely) `forget' some of the training samples.  More specifically, we modify the training objective so that the empirical distribution $P_n$ is allowed to be perturbed within a neighborhood of   distributions $Q$, where the neighborhood is defined by a chosen $f$-divergence (e.g., KL divergence) and a  size parameter $\delta>0$, leading to the new stochastic optimization problem
\begin{align}\label{eq:relaxed_optim}
\inf_\theta \inf_{Q: D_f(Q\|P_n)\leq \delta} E_{Q}[\mathcal{L}_\theta]\,.
\end{align}
Intuitively,  minimizing over $Q$ causes the method to search for a nearby distribution over  the training samples that reduces the expected loss. A model is then fit to that relaxed problem. Thus we refer to \eqref{eq:relaxed_optim} as a relaxation-optimization method.  The  relaxation allows the method to  ignore problematic samples, i.e., to `forget' the samples that have label errors.   We emphasize that it is the min-min optimization over a model neighborhood \eqref{eq:relaxed_optim} (i.e., relaxation), as opposed to a min-max formulation as used in, e.g., adversarial robustness \cite{madry2018towardsrobustoptimization}, that gives our method the ability to identify and forget incorrectly labeled samples; in contrast to \eqref{eq:relaxed_optim}, a min-max formulation would increase the effect of  mislabeled samples on the training. This insight is one of the key innovations of our work. See Theorem \ref{thm:exact_sol_main} below for a rigorous analysis of the ability of  \eqref{eq:relaxed_optim} to `forget' samples with label errors.

The choice of $\delta>0$ in \eqref{eq:relaxed_optim}  controls the size of the relaxation neighborhood and can be viewed as the method's `forgetting' budget.   We specifically work with model neighborhoods defined by $f$-divergences for two reasons: 1) $D_f(Q\|P_n)\leq \delta$ implies $Q\ll P_n$ and so $Q$ will   reweight the original training samples, without otherwise perturbing them. 2) $f$-divergences satisfy several important properties which enable an efficient numerical implementation; for details, see Section \ref{sec:duality}. The implementation takes the form of an adversarial method, where sample weights are dynamically updated during training in a principled manner, allowing the method to determine the optimal subset of samples to ignore.

%%Our work shows that there exists a significant room to improve learning in the presence of noisy lables without relying on clean data or any effort towards label correction.

Based on the structure of our method's implementation (see Algorithm \ref{alg:ANTIDOTE}) and its computational cost, we argue that \ant~ is  best compared to other methods that employ a modified loss function.  Extensive evaluations on various levels of symmetric, asymmetric, human annotated, and real-world label noise across five datasets show that \ant~ achieves a significantly higher performance compared to its latest counterpart proposed in \cite{wangepsilon_2024_neurips} and other well-established benchmark methods in the field. The implementation of \ant~ is available at \url{https://github.com/star-ailab/ANTIDOTE}.

In summary, our main contributions are as follows.

{
~- We introduce a theoretically sound information-theoretic problem formulation \eqref{eq:relaxed_optim}   in terms of a relaxation over an $f$-divergence neighborhood, which has the effect of detecting and reducing the impact of noisy labels on the training. We prove in Theorem \ref{thm:exact_sol_main} that, under appropriate assumptions, its solution produces the true (non-noisy) labels.
    
~- Using convex duality, in Theorem \ref{thm:duality} we provide a computationally efficient  solution method for \eqref{eq:relaxed_optim},  in the form of a modified loss function paired with an adversarial low-dimensional (1-D or 2-D)   convex optimization problem, which can be easily integrated into any supervised deep learning model. Due to its low-dimension and simple  nature, this auxiliary convex optimization adds very little to the overall computational cost of the method.  Thus, \ant~ enjoys a time complexity that is very close to that of the standard cross entropy loss  (see Table \ref{tab:time}).

}

\section{Related Work}\label{sec:related_work}

One line of  work  for addressing noisy labels \citep{zhou2024l2b, ren2018learning_reweight_ICML_Baseline_MR,xia2023combating_sampleselection, sohn2020fixmatch_pseudolabel} studies robust losses combined with refinement techniques that require clean labels. Although this learning paradigm can lead to high model performance, clean labels are not always known a priori in practice (e.g., in training data poisoning attack scenarios). To remove reliance on clean data, another stream of recent work performs label correction procedures \citep{chang2023csot_labelcorrection_optimaltransport} via pseudo-label prediction techniques such as MixUP \citep{sohn2020fixmatch_pseudolabel,zhang2018mixup}. Other relevant works in this stream have shown promising results by performing these techniques on self-supervised pre-trained models \citep{zheltonozhskii2022contrasttodivide_c2d, Li2020DivideMix}. As these approaches combine multiple techniques, they often perform well at the cost of a significantly increased training time \citep{zhou2024l2b}.

Another major stream of related work focuses on designing new learning objectives that are specially tailored to operate in the presence of noisy labels. These studies include Focal Loss (FL) \cite{ross2017focal_loss}, Mean Absolute Error (MAE) loss \cite{ghosh2017robust_MAE_loss}, Generalized Cross Entropy (GCE) loss \cite{zhang2018generalized_GCE_loss}, Symmetric Cross Entropy (SCE) loss, Reverse Cross Entropy (RCE) loss \citep{wang2019symmetric_RCE_Loss}, \cite{wang2019symmetric_SCE_loss}, Active Passive Loss (APL) \cite{ma2020normalized_APL_loss}, Asymmetric Unhinged Loss (AUL) \cite{zhou2021asymmetric_loss}, Asymmetric
Exponential Loss (AEL) \cite{zhou2021asymmetric_loss}, Asymmetric GCE (AGCE) \cite{zhou2021asymmetric_loss}, LDR-KL loss \cite{zhu2023label_LDR_KL_ICML},
LogitClip (LC) loss \cite{wei2023mitigating_LC_loss}, and the recently proposed $\epsilon$-softmax loss \cite{wangepsilon_2024_neurips}. Building such objectives often revolves around an effective way to reweight samples based on their corresponding loss values, with the aim of reducing the effect of noisy labels on the model \citep{zhou2021asymmetric_loss, zhu2023label_LDR_KL_ICML,wangepsilon_2024_neurips}.   { Further approaches are based on  estimating the noise transition matrix, which  can then be used for resampling or reweighting \cite{natarajan2013learning,liu2015classification,patrini2017making,xia2019anchor,baedirichlet}.}

{ Conceptually, the relaxation mechanism that defines our method \eqref{eq:relaxed_optim}  is quite different from the above approaches. While relaxation-optimization over $f$-divergence neighborhoods does constitute a type of reweighting, it is of a very different character than the previously mentioned transition-matrix based methods.  On the other hand, operationally   our method can be viewed as a modified loss, though the interpretation of  our relaxation-optimization formulation is quite different from the  robust-loss   perspective.   We note that prior works have, in other forms,  used $f$-divergences  as tools to address label noise. The work of \cite{wei2021optimizing_F-Div} proposed maximizing $f$-divergence mutual information in place of standard CE-loss minimization and showed that, under certain assumptions, or by including a correction term which can be estimated from the data, this produces a robust classifier in a certain sense.  This approach is quite different from \ant, which uses $f$-divergences neighborhoods to construct the relaxed problem \eqref{eq:relaxed_optim}, and which can be interpreted as detecting and filtering out  the samples with noisy labels (see the discussion in Section \ref{sec:lambda_rho}). Also, the recent  approach   of \cite{chen2024mitigating}  employed a different min-min formulation, involving a special case of $f$-divergence soft-constraint (specifically, total variation), in contrast to the general $f$-divergence-neighborhood (hard) constraint in \eqref{eq:relaxed_optim}.  We note that it is the hard constraint that allows us to prove in Theorem \ref{thm:exact_sol_main} that the true labels provide the unique solution to the relaxed optimization problem \eqref{eq:relaxed_optim} under appropriate assumptions.  Another innovation of our method is Theorem \ref{thm:duality}, which derives a dual representation of \eqref{eq:relaxed_optim} where the optimization over $Q$ is replaced by a low-dimensional (1-D or 2-D) convex problem. From an implementation and computational complexity perspective,  this places  \ant~ most clearly within the realm of the modified learning objective methods such as the recent $\epsilon$-softmax loss method \cite{wangepsilon_2024_neurips}, despite the conceptual differences between them; see Table \ref{tab:time}.  Specifically, as seen in   \eqref{eq:reformulation_final},
\eqref{eq:reformulation_final_KL}, and Algorithm \ref{alg:ANTIDOTE} below, in practice our method amounts to an appropriate modification of the loss function together with the aforementioned inexpensive 1-D or 2-D convex optimization problem.    } 

{ In our results in Section \ref{sec:results}, we primarily focus on comparing the performance of our method with current leading modified losses which, as discussed above, are similar to \ant~ from the operational and computational cost perspectives. Additional comparison with transition matrix methods can be found in Appendix \ref{app:transition_matrix_comp}.  In line with recent studies\cite{zhou2021asymmetric_loss, wangepsilon_2024_neurips}, we also note that  modified learning objectives, including ours, can also be used in combination with  previously mentioned enhancement techniques, e.g., label correction, pseudo label prediction, Mixup label interpolation, semisupervised learning, or simultaneously training two neural nets such as Co-Teaching/Co-Teaching$+$\citep{yu2019does_COTeachingPlus, han2020robust_NeurIPS_COTeaching}, DivideMix \citep{Li2020DivideMix}, ELR$+$ \citep{liu2020early_ELR}, SOP$+$\cite{liu2022robust_SOP},   C2D \cite{zheltonozhskii2022contrasttodivide_c2d}, and CSOT\citep{chang2023csot_labelcorrection_optimaltransport}; see our additional evaluations in Section \ref{sec:add_ons} for details along this direction.

\section{Theoretical Results}\label{sec:theory}
In this section we present theoretical results that support the effectiveness of the relaxed problem \eqref{eq:relaxed_optim} and also derive a mathematically equivalent reformulation of the optimization problem \eqref{eq:relaxed_optim} that is more efficient for computational purposes. First we recall the definition and needed properties of $f$-divergences, which are used to define the model neighborhoods in \eqref{eq:relaxed_optim}.

\subsection{Background on $f$-Divergences}
Introduced in \cite{ali1966general,csiszar1967information}, the $f$-divergences quantify the discrepancy between probability distributions, $Q$ and $P$, by measuring the size of the likelihood ratio $dQ/dP$ in an appropriate sense. More specifically, given a convex function $f$ on the real line with $f(1)=0$, the corresponding $f$-divergence is defined by
\begin{align}\label{eq:f_div_def}
    D_f(Q\|P)\coloneqq E_P[f(dQ/dP)]\,.
\end{align}
When the likelihood ratio does not exist, i.e.,  when $Q\not\ll P$, then the appropriate definition for our purposes is to let $D_f(Q\|P)\coloneqq\infty$. The most well-known   $f$-divergence is the KL divergence, which corresponds to $f_{\mathrm{KL}}(t)\coloneqq t\log(t)$. A key  property of the $f$-divergences that we use below is the convexity of $Q\mapsto D_f(Q\|P)$; this follows from the variational representation of $f$-divergences \cite{Broniatowski,Nguyen_Full_2010,birrell2022f}.  Further discussions of the properties of  $f$-divergences can be found in, e.g., \cite{LieseVajda,birrell2022f}.

\subsection{Exact Solution to the Relaxed Problem}
In this section we present a theorem showing that, under appropriate assumptions, the noise-free labels provide a (unique) solution to the relaxed problem \eqref{eq:relaxed_optim} with noisy labels.  This provides theoretical backing to the claim that \ant~ is an effective method for learning in the presence of label noise.

We work with distributions over samples with exact and noisy labels defined as follows. Given a probability distribution $P_\mathcal{X}$ on the sample space $\mathcal{X}$  with deterministic labels defined by the measurable map $h_*:\mathcal{X}\to \mathcal{Y}$ (where $\mathcal{Y}$ is a finite set of labels), we define $P_{*}$ to be the corresponding distribution on  $\mathcal{X}\times\mathcal{Y}$, the space of sample/label pairs, i.e.,
\begin{align}\label{eq:P_star_def_main}
P_{*}\coloneqq\delta_{h_*(x)}(dy) P_{\mathcal{X}}(dx)\,.
\end{align}
Given a measurable  sample-dependent transition matrix, $T^x_{y,\tilde{y}}$ (i.e., the probability of the sample $x$ having its label, $y$,  flipped to the corrupted label $\tilde{y}$), we  define the corresponding distribution on samples and corrupted labels  by
\begin{align}\label{eq:P_T_def_main}
 P_{T}\coloneqq \sum_{\tilde{y}\in\mathcal{Y}} T^x_{h_*(x),\tilde y}\delta_{\tilde{y}}(dy) P_{\mathcal{X}}(d{x})\,.  
\end{align}

The following theorem gives conditions guaranteeing  that the relaxed problem \eqref{eq:relaxed_optim}, with $P_n$ replaced by $P_T$ (i.e., working with the abstract distribution \eqref{eq:P_T_def_main} as opposed to the empirical distribution of a set of samples), is uniquely solved by  $h_*$, the noise-free labels. For the proof, see Theorem \ref{thm:h_star_unique_sol_app} in Appendix \ref{app:thm_optimizer}. There we also prove under much weaker assumptions that $h_*$ provides a (possibly non-unique) solution; see Theorem \ref{thm:exact_labels_solve_relaxed_problem}.
\begin{theorem}\label{thm:exact_sol_main}
    Let $\mathcal{X}$ be a complete separable metric space with its Borel sigma algebra and let $P_{\mathcal{X}}$, $h_*$, $P_*$,  $P_T$ be as in  \eqref{eq:P_star_def_main}-\eqref{eq:P_T_def_main}. Suppose the following.
\begin{enumerate}
\item We have $r\in(0,1)$ such that  $T^x_{h^*(x),h^*(x)}=1-r$ for all $x\in\mathcal{X}$ and
\begin{align}
   T^x_{h_*(x),y}<1-r\, \text{ for all $x\in\mathcal{X},y\in\mathcal{Y}$  that satisfy $y\neq h_*(x)$}\,.
\end{align}
Note that the latter condition automatically follows from the former if $r<1/2$.
\item We have a classifier family $h_\theta:\mathcal{X}\to\mathcal{P}(\mathcal{Y})$, $\theta\in \Theta$,  where $\mathcal{P}(\mathcal{Y})$ denotes the space of probability vectors over the class labels. We also suppose that there exists $\theta_*\in\Theta$ such that $h_{\theta_*}=h_*$ (here and below we identify  each label with its corresponding one-hot vector).
\item We have a loss function ${L}:\mathcal{P}(\mathcal{Y})\times \mathcal{Y}\to [0,\infty]$ such that ${L}(p,y)=0$ if and only if $p=y$. 
\item Equip $\mathcal{Y}$ with the discrete metric, $d_{\mathcal{Y}}(y_1,y_2)\coloneqq 1_{y_1\neq y_2}$ and assume that $\mathcal{L}_\theta(x,y)\coloneqq L(h_\theta(x),y)$ is lower semicontinuous  on $\mathcal{X}\times\mathcal{Y}$ for all $\theta\in\Theta$.
\item We have  a strictly convex $f:(0,\infty)\to \mathbb{R}$ such that  $f(1)=0$, $f(0)\coloneqq \lim_{t\to 0^+} f(t)$ is finite, and  $f^*(y)$ is finite for all $y\in\mathbb{R}$ \,(e.g., $f_{\mathrm{KL}}$ or $f_\alpha$ for $\alpha>1$).
\end{enumerate}
Define $\delta\coloneqq rf(0)+(1-r)f(1/(1-r))$.  Then for all
\begin{align}
 \tilde{\theta} \in\argmin_{ \theta\in\Theta}\inf_{Q: D_f(Q\|P_T)\leq \delta} E_{Q}[\mathcal{L}_\theta]
\end{align}
we have $h_{\tilde{\theta}}=h_*$ almost surely under $P_{\mathcal{X}}$.
\end{theorem}

Note that while, a priori, the classifiers are valued in $\mathcal{P}(\mathcal{Y})$, Theorem \ref{thm:exact_sol_main} implies that the optimal solution to the relaxed problem is valued in $\mathcal{Y}$, i.e.,  it predicts the true class labels with complete confidence. This is not the case for solutions to standard training under label noise.  In this sense, the \ant~ method filters out the noisy labels; they have no impact the optimal solution.  This view is further reinforced in Section \ref{sec:lambda_rho} where we explore the behavior of \ant~ during training.

\subsection{Reformulation via Convex Duality}\label{sec:duality}
Next we use convex duality theory to derive a more computationally efficient form of our method \eqref{eq:relaxed_optim}.  Here we consider a more general version, which  allows for an additional soft-constraint penalty term $\kappa D_f(Q\|P_n)$, $\kappa\geq 0$, to be added to the objective in \eqref{eq:relaxed_optim}.  In practice, we find that using a small $\kappa>0$ improves performance and prevents numerical issues. Its use can be  motivated mathematically by the fact that it makes the objective strictly convex in $Q$, whenever $f$ is strictly convex, and thus ensure uniqueness of the minimizing $Q$.

 To begin the derivation, note that the optimization over $Q$ in \eqref{eq:relaxed_optim} has the form of a convex optimization problem. Specifically, the constraint $Q\mapsto D_f(Q\|P_n)$ is convex and the objective $Q\mapsto E_Q[\mathcal{L}_\theta]+\kappa D_f(Q\|P_n)$ (including the added soft constraint) is also convex in $Q$.  We can therefore rewrite the optimization as an equivalent saddle point problem using  convex duality theory, e.g., Theorem 3.11.2  in \cite{ponstein2004approaches}, which yields
\begin{align}
&\inf_{Q: D_f(Q\|P_n)\leq \delta}\{E_Q[\mathcal{L}_\theta]+\kappa D_f(Q\|P_n)\}\\
=&\sup_{\lambda>0}\{-\lambda \delta+\inf_{Q }\{E_Q[\mathcal{L}_\theta]+(\lambda+\kappa) D_f(Q\|P_n)\}\}\,.\notag
\end{align}
The unconstrained maximization over $Q$ can then be evaluated using the Gibbs variational formula for $f$-divergences, see Theorem 4.2 in \cite{BenTal2007}, yielding
\begin{align}
  \inf_{Q }\{E_Q[\mathcal{L}_\theta]+(\lambda+\kappa) D_f(Q\|P_n)\}
  =&-(\lambda+\kappa)\sup_{Q }\{E_Q[-\mathcal{L}_\theta/(\lambda+\kappa)]- D_f(Q\|P_n)\}\\
  =&-(\lambda+\kappa)\inf_{\nu\in\mathbb{R} }\{\nu+E_{P_n}[f^*(-\mathcal{L}_\theta/(\lambda+\kappa)-\nu)]\}\notag\,.
\end{align}
This formula is another reason for defining the optimization neighborhood in terms of $f$-divergences.  After changing variables to $\rho=(\lambda+\kappa)\nu$, the result is the following  problem reformulation
\begin{align}\label{eq:reformulation_final}
&\inf_{\theta\in\Theta}\inf_{Q: D_f(Q\|P_n)\leq \delta}\{E_Q[\mathcal{L}_\theta]+\kappa D_f(Q\|P_n)\}\\
=&\inf_{\theta\in\Theta}\sup_{\substack{\lambda>0,\\\rho\in\mathbb{R}} }\!\left\{\!-\lambda \delta-\rho-(\lambda+\kappa)E_{P_n}\!\!\left[f^*\!\!\left(-\frac{\mathcal{L}_\theta+\rho}{\lambda+\kappa}\right)\!\right]\right\}\,.\notag
\end{align}
As a result of minimizing over the relaxation neighborhood $\{Q:D_f(Q\|P_n)\leq \delta\}$, the objective now involves  the Legendre transform of $f$, defined by $f^*(y)=\sup_{x>0}\{yx-f(x)\}$. It  also includes  two additional  parameters, $\lambda$ and $\rho$, which appear in an adversarial (i.e., min-max) form; however, we emphasize that here the inner maximization does not involve a second neural network (NN), in contrast to dual-NN methods such as GANs. { More specifically, the inner maximization in our method is a computationally inexpensive low-dimensional convex problem; for more implementation details, including pseudocode, see Appendix \ref{app:pseudocode}}.   Intuitively, the effect of $f^*$ can be viewed as reweighting the samples, while $\lambda$ and $\rho$ dynamically influence the amount of reweighting; we will further discuss their significance in Section \ref{sec:lambda_rho} below; however, we again emphasize that the reweighting in our method is very different from reweighting via transition-matrix approaches. The form of $f^*$ can be determined explicitly in a number of important cases (e.g., KL and $\alpha$-divergences), and thus $f^*$ does not contribute significantly to the computational cost.   In the special case of the KL divergence one can further evaluate the optimization over $\rho$ to obtain    
\begin{align}\label{eq:reformulation_final_KL}
&\inf_{\theta\in\Theta}\inf_{Q:\mathrm{KL}(Q\|P_n)\leq \delta} \{E_{Q}[\mathcal{L}_\theta]+\kappa \mathrm{KL}(Q\|P_n)\} \\
=&\inf_{\theta\in\Theta}\sup_{\lambda>0}\!\left\{-\lambda\delta-(\lambda+\kappa)\log E_{P_n}[\exp(-\mathcal{L}_\theta/(\lambda+\kappa))]\right\}\,.\notag
    \end{align}

    Performing alternating gradient descent/ascent on the objective on the right-hand sides of \eqref{eq:reformulation_final} or \eqref{eq:reformulation_final_KL} is only slightly more computationally expensive than SGD for the standard learning problem \eqref{eq:relaxed_optim}. Moreover, recalling the convexity of $f^*$, of the cumulant generating function, and also the fact that the perspective of a convex function is convex, we  see that the objective function on the right-hand side of \eqref{eq:reformulation_final} is concave in $(\lambda,\rho)$ and the objective  on the right-hand side of \eqref{eq:reformulation_final_KL} is concave in $\lambda$.  Thus the inner  optimizations are  low-dimensional convex problems. In lieu of SGD,  we find that applying specialized optimization methods to the important parameters $(\lambda,\rho)$ can result in significant performance benefits with little added computational cost; see Appendix \ref{app:pseudocode} for further discussion.  
    
    In summary, we have  the following result; see Appendix \ref{app:duality_proof} for a detailed proof.
\begin{theorem}\label{thm:duality}
Let $P_n$ be the empirical distribution of the samples $(x_i,y_i)$, $i=1,...,n$, and suppose:
\begin{enumerate}
    \item We have $0\leq a<1<b\leq \infty$ and a convex $f:(a,b)\to\mathbb{R}$ that satisfies $f(1)=0$.
    \item  We have a real-valued loss $\mathcal{L}_\theta(x,y)$, depending on parameters $\theta\in\Theta$.
\end{enumerate}
Then for all $\delta>0$, $\kappa\geq 0$ we have the equality \eqref{eq:reformulation_final} and in the KL case, we have the further simplification \eqref{eq:reformulation_final_KL}.  Moreover, the objective function on the right-hand side of \eqref{eq:reformulation_final} is concave in $(\lambda,\rho)$ and the objective  on the right-hand side of \eqref{eq:reformulation_final_KL} is concave in $\lambda$.
\end{theorem}
Mathematically, the proof of Theorem \ref{eq:reformulation_final_KL}  is  similar to that of the distributionally robust optimization (DRO) result for $f$-divergence neighborhoods found in Theorem 5.1 of \cite{Javid2012}.  However, the change from maximization over the model neighborhood $\{Q:D_f(Q\|P_n)\leq \delta\}$ (as in DRO) to the minimization  used here has significant practical consequences for the model training, i.e., when combined with the minimization over $\theta$. Intuitively, by minimizing over the model neighborhood we reduce the weights of samples that have a large loss at each step in the training, motivated by the heuristic that those samples are more likely to be incorrectly labeled.  This is in contrast to DRO methods, e.g.,  as used for adversarial robustness \cite{madry2018towardsrobustoptimization,zhang2019theoreticallytradeOff,wang2019improving_mart,bui_UDR_2022unified,ARMOR}, which search for small perturbations to the training samples that greatly increase the loss.  In the next two subsections we further analyze several aspects of  \eqref{eq:reformulation_final} and \eqref{eq:reformulation_final_KL}, providing additional insight into why this modification should be expected to improve performance when learning with noisy labels.

\subsection{Interpretation of the  Adversarial Parameters}\label{sec:lambda_rho}
To understand the importance of the $\lambda$ and $\rho$ parameters in \eqref{eq:reformulation_final} we study the form of the objective and its optimization in two important  cases, that of $\alpha$-divergences and the KL-divergence.  These illustrate two qualitatively different types of behavior.

The $\alpha$-divergences correspond to $f_\alpha(t)=      \frac{t^\alpha-1}{\alpha(\alpha-1)}$, $\alpha>1$, which have Legendre transform  
\begin{align}\label{eq:f_alpha_star}
f_\alpha^*(t)=&\alpha^{-1}(\alpha-1)^{\alpha/(\alpha-1)}\text{ReLU}(t)^{\alpha/(\alpha-1)}+(\alpha(\alpha-1))^{-1}\,.
\end{align}
The key feature of $f_\alpha^*$ is that it is constant on $(-\infty,0)$, due to the $\text{ReLU}$ function, and so samples  $(x_i,y_i)$ will be `forgotten' if $-(\mathcal{L}_\theta(x_i,y_i)+\rho)/\lambda<0$, i.e., if $\mathcal{L}_\theta(x_i,y_i)>-\rho$, because they will have no effect on the gradient (with respect to $\theta$) of the objective in \eqref{eq:reformulation_final}. The parameter $\rho$ dynamically sets the forgetfulness threshold, so that samples that are forgotten in one SGD step may be `remembered' in subsequent steps.

In the case of KL-neighborhoods, there is no hard forgetting-threshold. Rather, the sample weights are exponentially suppressed based on the corresponding loss values.  This can be seen via the following calculation. The gradient of the objective in \eqref{eq:reformulation_final_KL} on a 
minibatch of size $B$ can be written
\begin{align}\label{eq:KL_gradient}
    &\nabla_\theta\left[-\lambda \delta-\lambda\log \left(\frac{1}{B}\sum_{i=1}^B\exp(-\mathcal{L}_\theta(x_i,y_i)/\lambda)\right)\right]
    =\frac{1}{B}\sum_{i=1}^Bw_i\nabla_\theta \mathcal{L}_\theta(x_i,y_i)\,,\\
    &w_i\coloneqq\frac{\exp(-\mathcal{L}_\theta(x_i,y_i)/\lambda) }{ \frac{1}{B}\sum_{j=1}^B\exp(-\mathcal{L}_\theta(x_j,y_j)/\lambda)}\,.\label{eq:KL_weights}
\end{align}
From this we see that the gradient of the baseline loss has been modified by the weights $w_i$.
The $i$'th weight will be very small when the loss at the $i$'th sample is much larger than loss at the best-fit sample in the minibatch. The factor $\lambda$ determines how quickly the weights decay as the difference in loss values increases, with this rate dynamically changing throughout the algorithm based on the inner optimization over $\lambda$ in \eqref{eq:reformulation_final_KL}.   If the parameter $\theta$ is (approximately) a local minimizer of the loss at the $i$'th sample   then $\nabla_\theta \mathcal{L}_\theta(x_i,y_i)$ will  be small and so the $i$'th term in the sum on the right-hand side of \eqref{eq:KL_gradient} will also be small.  Therefore, intuitively one can think of the gradient update as focusing the most attention on the samples in the minibatch that are moderately out of fit, in  that $\mathcal{L}_\theta(x_i,y_i)$ is not too large   but also where $\nabla_\theta\mathcal{L}_\theta(x_i,y_i)$ is not too small.  Similar remarks apply to the $\alpha$-divergence case for samples that are above the forgetfulness threshold, as determined  dynamically by $\rho$, with the $\lambda$ parameter again dynamically determining the rate at which the weights are suppressed.  

The behavior discussed here is in contrast with standard training, which is prone to overly focusing on outliers; that can be very problematic when those outliers are actually incorrectly labeled samples.  Using a standard DRO-based loss, i.e., maximizing over $\{Q:D_f(Q\|P_n)\leq \delta\}$ instead of minimizing,  only exacerbates this focus on outliers. Specifically, a min-max based loss applied to KL would involve weights  of the form \eqref{eq:KL_weights} except with the signs of the exponents all reversed; therefore it would focus the training  on the samples with the largest loss. While that behavior is reasonable in the case of  adversarial robustness (where   min-max DRO methods are widely used), in the presence of noisy labels such a focus on outliers is counterproductive. Thus, based on the above calculations, we argue  that the  min-min optimization over a model neighborhood \eqref{eq:relaxed_optim} is  more appropriate for constructing a robust objective to learn from noisy labels.   { We also point out the LDR-KL method \cite{zhu2023label_LDR_KL_ICML}, which used a DRO approach at the level of the loss function, with the sample fixed, wherein the classes are reweighted to encourage the model to be less-certain about its predictions in order to avoid overly fitting the mislabeled samples. This is a quite different approach from our relaxation-optimization method, which   detects and filters out mislabeled samples from influencing the loss, rather than acting on a per-sample basis like in LDR-KL; comparisons on experiments can be found in Tables \ref{tab:benchmark} and \ref{tab:real}.
}

\subsection{Interpretation of the $f$-Divergence Neighborhood Size}\label{sec:delta_interp}

In this section we provide an operational interpretation of the $f$-divergence neighborhood size, $\delta$.  Specifically,  $\delta$ is related to the maximum fraction of samples that can be forgotten, denoted $r_{max}$, via
\begin{align}\label{eq:delta_rmax}
    \delta=r_{max}f(0) +(1-r_{max}) f(1/(1-r_{max}))\,.
\end{align}
A derivation can be found in Appendix \ref{app:nbhd_interp}.
In particular, in the KL case this simplifies to $\delta=\log(1/(1-r_{max}))$. We emphasize that \eqref{eq:delta_rmax} relates $\delta$ to the maximum forgetfulness fraction, which occurs when all other samples are equally weighted.  If the neighborhood is to contain distributions that forget a fraction $r$ of samples and also assign  the remaining samples unequal weights then  the corresponding $\delta$ must be larger.   Therefore, even if one knows the true fraction of samples that have incorrect labels, the choice of $\delta$ that leads to optimal performance will generally differ from what is suggested by \eqref{eq:delta_rmax}.  In practice, we find that \eqref{eq:delta_rmax} gives a good starting point for choosing $\delta$, which we view as a tunable hyperparameter for our method. { Moreover, based on the formula \eqref{eq:delta_rmax}, we propose a variant of \eqref{eq:reformulation_final} that adaptively  selects the value of $\delta$ via a penalized optimization procedure; see Appendix \ref{app:delta_penalized_opt} for details.}

\section{Experiments}\label{sec:results}
We compare the performance of \ant~ with multiple benchmark methods, including standard cross entropy loss (CE), FL \cite{ross2017focal_loss}, MAE loss \cite{ghosh2017robust_MAE_loss}, GCE \cite{zhang2018generalized_GCE_loss}, SCE \cite{wang2019symmetric_SCE_loss}, RCE \citep{wang2019symmetric_RCE_Loss}, APL \cite{ma2020normalized_APL_loss}, AUL \cite{zhou2021asymmetric_loss}, AEL \cite{zhou2021asymmetric_loss}, AGCE \cite{zhou2021asymmetric_loss}, Negative Learning for Noisy Labels (NLNL) \cite{kim2019nlnl_loss},  LDR-KL \cite{zhu2023label_LDR_KL_ICML}, 
LogitClip \cite{wei2023mitigating_LC_loss}, and $\epsilon$-softmax loss \cite{wangepsilon_2024_neurips}. For MAE, RCE, and asymmetric losses (AEL, AGCE, and AUL) we also considered their combination with normalized CE.   We followed the experimental settings in \citep{wangepsilon_2024_neurips, zhou2021asymmetric_loss, ma2020normalized_normalized}. To ensure reproducibility, we integrated our implementation into the code provided by \cite{wangepsilon_2024_neurips, zhou2021asymmetric_loss}.  Detailed parameter settings for \ant~ and all the benchmark methods appear in Appendix~\ref{app:param}. { Parameter sensitivity and ablation analysis of \ant~ is provided in Appendix \ref{app:sensitivity}.} \ant's implementation  is available online at \url{https://github.com/star-ailab/ANTIDOTE}.

\subsection {Empirical Validity of \ant~ Loss}\label{sec:histogram}

Before conducting benchmark experiments, we perform a preliminary but critical analysis to evaluate whether \ant~  is capable of distinguishing between noisy and clean samples as training proceeds. This experiment on CIFAR-10 serves as a sanity check on the validity of \ant~ before any extensive evaluations. We perform this test using the $\alpha$-divergence variant of our method, as its hard-`forgetting' threshold makes the interpretation clearer; see the discussion in Section \ref{sec:lambda_rho}. Figure~\ref{fig:losses_cropped} shows the histograms of \ant~ losses for different noise ratios at different stages of training. Recalling that, in the $\alpha$-divergence case, samples are forgotten, i.e., do not influence the training, when $-(\mathcal{L}+\rho)<0$, we plot the histograms with the $x$-axis being $-(\mathcal{L}+\rho)$. Figures~\ref{fig:losses_cropped}.(a), (b), and (c) shows the  histograms for the 50\% symmetric noise ratio at 50 (a), 100 (b), and 200 (c) epochs. The losses corresponding to clean samples are shown in blue and the ones associated with noisy samples are depicted in orange.
\begin{figure*}[h!]
\vspace{-5pt}
    \centering
    \includegraphics[width=1.0\linewidth]{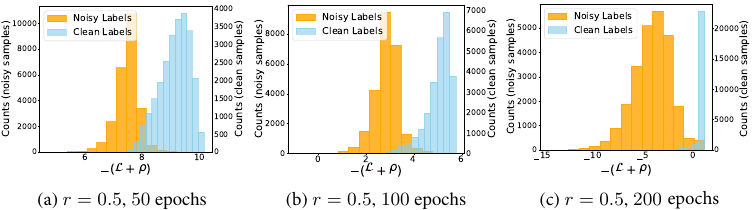}
    \caption{\ant's loss for different noise ratios during training}
    \label{fig:losses_cropped} \vspace{-4mm}
\end{figure*}

From the results shown in Figure~\ref{fig:losses_cropped}, it is evident that during training the clean loss distribution is to the right side of the noisy loss distribution, implying that \ant~ is able to separate the noisy samples from the clean ones for each of these  noise levels. As the training proceeds from 50 epochs to 200 epochs this separation becomes clearer, with the losses of the samples with noisy labels largely moving below the forgetting threshold at $-(\mathcal{L}+\rho)=0$. This suggest that \ant~ effectively learns to distinguish between the clean and noisy labeled samples and is correctly able to determine that the samples with noisy labels should not influence the training of the model.  We emphasize that the training algorithm was not provided with any (full or partial) information regarding which samples had the correct labels.  

\begin{table*}[h!]
\small
\centering
\setlength\tabcolsep{1.1 pt}
\caption{Last epoch test accuracies (\%) of different methods on CIFAR-10 and CIFAR-100 at various levels of symmetric and asymmetric noise. All results are reported over 3 random runs (mean±std). The results for our \ant~ method all used the KL variant \eqref{eq:reformulation_final_KL}.}
\label{tab:benchmark}
\begin{tabular}{ll|llll|llll}
\toprule
CIFAR-10 &
  \multicolumn{1}{c}{\multirow{2}{*}{Clean}} &
  \multicolumn{4}{c}{Symmetric Noise Rate} &
  \multicolumn{4}{c}{Asymmetric Noise Rate} \\ %%\cline{1-1} \cline{3-10} 
Loss &
  \multicolumn{1}{c}{} &
  \multicolumn{1}{c}{0.2} &
  \multicolumn{1}{c}{0.4} &
  \multicolumn{1}{c}{0.6} &
  \multicolumn{1}{c}{0.8} &
  \multicolumn{1}{c}{0.1} &
  \multicolumn{1}{c}{0.2} &
  \multicolumn{1}{c}{0.3} &
  \multicolumn{1}{c}{0.4} \\ \hline
CE &
  90.50\fontsize{7pt}{7pt}\selectfont±0.35 &
  \multicolumn{1}{l}{75.47\fontsize{7pt}{7pt}\selectfont±0.27} &
  \multicolumn{1}{l}{58.46\fontsize{7pt}{7pt}\selectfont±0.21} &
  \multicolumn{1}{l}{39.16\fontsize{7pt}{7pt}\selectfont±0.50} &
  18.95\fontsize{7pt}{7pt}\selectfont±0.38 &
  \multicolumn{1}{l}{86.98\fontsize{7pt}{7pt}\selectfont±0.31} &
  \multicolumn{1}{l}{83.82\fontsize{7pt}{7pt}\selectfont±0.04} &
  \multicolumn{1}{l}{79.35\fontsize{7pt}{7pt}\selectfont±0.66} &
  75.28\fontsize{7pt}{7pt}\selectfont±0.58 \\
FL &
  89.70\fontsize{7pt}{7pt}\selectfont±0.24 &
  \multicolumn{1}{l}{74.50\fontsize{7pt}{7pt}\selectfont±0.18} &
  \multicolumn{1}{l}{58.23\fontsize{7pt}{7pt}\selectfont±0.40} &
  \multicolumn{1}{l}{38.69\fontsize{7pt}{7pt}\selectfont±0.06} &
  19.47\fontsize{7pt}{7pt}\selectfont±0.74 &
  \multicolumn{1}{l}{86.64\fontsize{7pt}{7pt}\selectfont±0.12} &
  \multicolumn{1}{l}{83.08\fontsize{7pt}{7pt}\selectfont±0.07} &
  \multicolumn{1}{l}{79.34\fontsize{7pt}{7pt}\selectfont±0.30} &
  74.68\fontsize{7pt}{7pt}\selectfont±0.31 \\ 
GCE &
  89.42\fontsize{7pt}{7pt}\selectfont±0.21 &
  \multicolumn{1}{l}{86.87\fontsize{7pt}{7pt}\selectfont±0.06} &
  \multicolumn{1}{l}{82.24\fontsize{7pt}{7pt}\selectfont±0.25} &
  \multicolumn{1}{l}{68.43\fontsize{7pt}{7pt}\selectfont±0.26} &
  25.82\fontsize{7pt}{7pt}\selectfont±1.03 &
  \multicolumn{1}{l}{88.43\fontsize{7pt}{7pt}\selectfont±0.20} &
  \multicolumn{1}{l}{86.17\fontsize{7pt}{7pt}\selectfont±0.29} &
  \multicolumn{1}{l}{80.72\fontsize{7pt}{7pt}\selectfont±0.42} &
  74.01\fontsize{7pt}{7pt}\selectfont±0.53 \\
NLNL &
  90.73\fontsize{7pt}{7pt}\selectfont±0.20 &
  \multicolumn{1}{l}{73.70\fontsize{7pt}{7pt}\selectfont±0.05} &
  \multicolumn{1}{l}{63.90\fontsize{7pt}{7pt}\selectfont±0.44} &
  \multicolumn{1}{l}{50.68\fontsize{7pt}{7pt}\selectfont±0.47} &
  29.53\fontsize{7pt}{7pt}\selectfont±1.55 &
  \multicolumn{1}{l}{88.5\fontsize{7pt}{7pt}\selectfont±0.25} &
  \multicolumn{1}{l}{84.74\fontsize{7pt}{7pt}\selectfont±0.08} &
  \multicolumn{1}{l}{81.26\fontsize{7pt}{7pt}\selectfont±0.43} &
  76.97\fontsize{7pt}{7pt}\selectfont±0.52 \\
SCE &
  91.30\fontsize{7pt}{7pt}\selectfont±0.08 &
  \multicolumn{1}{l}{87.58\fontsize{7pt}{7pt}\selectfont±0.05} &
  \multicolumn{1}{l}{79.47\fontsize{7pt}{7pt}\selectfont±0.48} &
  \multicolumn{1}{l}{59.14\fontsize{7pt}{7pt}\selectfont±0.07} &
  25.88\fontsize{7pt}{7pt}\selectfont±0.49 &
  \multicolumn{1}{l}{89.87\fontsize{7pt}{7pt}\selectfont±0.27} &
  \multicolumn{1}{l}{86.48\fontsize{7pt}{7pt}\selectfont±0.25} &
  \multicolumn{1}{l}{81.30\fontsize{7pt}{7pt}\selectfont±0.18} &
  74.99\fontsize{7pt}{7pt}\selectfont±0.16 \\
NCE+MAE &
  89.02\fontsize{7pt}{7pt}\selectfont±0.10 &
  \multicolumn{1}{l}{86.79\fontsize{7pt}{7pt}\selectfont±0.28} &
  \multicolumn{1}{l}{83.60\fontsize{7pt}{7pt}\selectfont±0.14} &
  \multicolumn{1}{l}{75.93\fontsize{7pt}{7pt}\selectfont±0.41} &
  46.96\fontsize{7pt}{7pt}\selectfont±0.67 &
  \multicolumn{1}{l}{88.03\fontsize{7pt}{7pt}\selectfont±0.27} &
  \multicolumn{1}{l}{85.53\fontsize{7pt}{7pt}\selectfont±0.08} &
  \multicolumn{1}{l}{81.10\fontsize{7pt}{7pt}\selectfont±0.52} &
  74.98\fontsize{7pt}{7pt}\selectfont±0.48 \\
NCE+RCE &
  91.03\fontsize{7pt}{7pt}\selectfont±0.28 &
  \multicolumn{1}{l}{88.41\fontsize{7pt}{7pt}\selectfont±0.24} &
  \multicolumn{1}{l}{85.13\fontsize{7pt}{7pt}\selectfont±0.56} &
  \multicolumn{1}{l}{79.20\fontsize{7pt}{7pt}\selectfont±0.06} &
  55.28\fontsize{7pt}{7pt}\selectfont±1.26 &
  \multicolumn{1}{l}{90.25\fontsize{7pt}{7pt}\selectfont±0.08} &
  \multicolumn{1}{l}{88.11\fontsize{7pt}{7pt}\selectfont±0.23} &
  \multicolumn{1}{l}{85.35\fontsize{7pt}{7pt}\selectfont±0.18} &
  79.43\fontsize{7pt}{7pt}\selectfont±0.21 \\
NFL+RCE &
  91.08\fontsize{7pt}{7pt}\selectfont±0.29 &
  \multicolumn{1}{l}{89.00\fontsize{7pt}{7pt}\selectfont±0.23} &
  \multicolumn{1}{l}{85.90\fontsize{7pt}{7pt}\selectfont±0.19} &
  \multicolumn{1}{l}{79.79\fontsize{7pt}{7pt}\selectfont±0.52} &
  55.47\fontsize{7pt}{7pt}\selectfont±2.73 &
  \multicolumn{1}{l}{89.99\fontsize{7pt}{7pt}\selectfont±0.35} &
  \multicolumn{1}{l}{88.33\fontsize{7pt}{7pt}\selectfont±0.26} &
  \multicolumn{1}{l}{85.27\fontsize{7pt}{7pt}\selectfont±0.13} &
  79.05\fontsize{7pt}{7pt}\selectfont±0.35 \\
NCE+AUL &
  91.06\fontsize{7pt}{7pt}\selectfont±0.24 &
  \multicolumn{1}{l}{89.11\fontsize{7pt}{7pt}\selectfont±0.07} &
  \multicolumn{1}{l}{85.79\fontsize{7pt}{7pt}\selectfont±0.16} &
  \multicolumn{1}{l}{79.57\fontsize{7pt}{7pt}\selectfont±0.21} &
  57.59\fontsize{7pt}{7pt}\selectfont±0.84 &
  \multicolumn{1}{l}{90.18\fontsize{7pt}{7pt}\selectfont±0.23} &
  \multicolumn{1}{l}{88.30\fontsize{7pt}{7pt}\selectfont±0.44} &
  \multicolumn{1}{l}{85.28\fontsize{7pt}{7pt}\selectfont±0.04} &
  79.14\fontsize{7pt}{7pt}\selectfont±0.36 \\
NCE+AGCE &
  91.13\fontsize{7pt}{7pt}\selectfont±0.11 &
  \multicolumn{1}{l}{89.00\fontsize{7pt}{7pt}\selectfont±0.29} &
  \multicolumn{1}{l}{85.91\fontsize{7pt}{7pt}\selectfont±0.15} &
  \multicolumn{1}{l}{80.36\fontsize{7pt}{7pt}\selectfont±0.36} &
  49.98\fontsize{7pt}{7pt}\selectfont±4.81 &
  \multicolumn{1}{l}{89.90\fontsize{7pt}{7pt}\selectfont±0.09} &
  \multicolumn{1}{l}{88.36\fontsize{7pt}{7pt}\selectfont±0.11} &
  \multicolumn{1}{l}{85.73\fontsize{7pt}{7pt}\selectfont±0.12} &
  79.28\fontsize{7pt}{7pt}\selectfont±0.37 \\
NCE+AEL &
  88.43\fontsize{7pt}{7pt}\selectfont±0.25 &
  \multicolumn{1}{l}{86.46\fontsize{7pt}{7pt}\selectfont±0.28} &
  \multicolumn{1}{l}{83.06\fontsize{7pt}{7pt}\selectfont±0.23} &
  \multicolumn{1}{l}{75.15\fontsize{7pt}{7pt}\selectfont±0.32} &
  43.22\fontsize{7pt}{7pt}\selectfont±0.46 &
  \multicolumn{1}{l}{87.59\fontsize{7pt}{7pt}\selectfont±0.38} &
  \multicolumn{1}{l}{85.98\fontsize{7pt}{7pt}\selectfont±0.14} &
  \multicolumn{1}{l}{82.87\fontsize{7pt}{7pt}\selectfont±0.16} &
  75.78\fontsize{7pt}{7pt}\selectfont±0.12 \\
LDR-KL &
  91.38\fontsize{7pt}{7pt}\selectfont±0.35 &
  \multicolumn{1}{l}{89.01\fontsize{7pt}{7pt}\selectfont±0.09} &
  \multicolumn{1}{l}{85.46\fontsize{7pt}{7pt}\selectfont±0.11} &
  \multicolumn{1}{l}{74.93\fontsize{7pt}{7pt}\selectfont±0.33} &
  34.78\fontsize{7pt}{7pt}\selectfont±0.67 &
  \multicolumn{1}{l}{90.24\fontsize{7pt}{7pt}\selectfont±0.18} &
  \multicolumn{1}{l}{88.38\fontsize{7pt}{7pt}\selectfont±0.02} &
  \multicolumn{1}{l}{85.03\fontsize{7pt}{7pt}\selectfont±0.16} &
  77.68\fontsize{7pt}{7pt}\selectfont±0.37 \\
CE+LC &
  90.06\fontsize{7pt}{7pt}\selectfont±0.41 &
  \multicolumn{1}{l}{85.66\fontsize{7pt}{7pt}\selectfont±0.32} &
  \multicolumn{1}{l}{79.18\fontsize{7pt}{7pt}\selectfont±0.57} &
  \multicolumn{1}{l}{53.87\fontsize{7pt}{7pt}\selectfont±0.57} &
  21.04\fontsize{7pt}{7pt}\selectfont±0.47&
  \multicolumn{1}{l}{87.99\fontsize{7pt}{7pt}\selectfont±0.06} &
  \multicolumn{1}{l}{84.01\fontsize{7pt}{7pt}\selectfont±0.01} &
  \multicolumn{1}{l}{79.71\fontsize{7pt}{7pt}\selectfont±0.51} &
  74.34\fontsize{7pt}{7pt}\selectfont±0.30 \\
CE${}_\epsilon$+MAE &
  91.40\fontsize{7pt}{7pt}\selectfont±0.12 &
  \multicolumn{1}{l}{89.29\fontsize{7pt}{7pt}\selectfont±0.10} &
  \multicolumn{1}{l}{85.93\fontsize{7pt}{7pt}\selectfont±0.19} &
  \multicolumn{1}{l}{79.52\fontsize{7pt}{7pt}\selectfont±0.14} &
  58.96\fontsize{7pt}{7pt}\selectfont±0.70 &
  \multicolumn{1}{l}{90.30\fontsize{7pt}{7pt}\selectfont±0.11} &
  \multicolumn{1}{l}{88.62\fontsize{7pt}{7pt}\selectfont±0.18} &
  \multicolumn{1}{l}{85.56\fontsize{7pt}{7pt}\selectfont±0.12} &
  78.91±\fontsize{7pt}{7pt}\selectfont0.25 \\ 
FL${}_\epsilon$+MAE &
  91.11\fontsize{7pt}{7pt}\selectfont±0.13 &
  \multicolumn{1}{l}{89.13\fontsize{7pt}{7pt}\selectfont±0.25} &
  \multicolumn{1}{l}{86.15\fontsize{7pt}{7pt}\selectfont±0.29} &
  \multicolumn{1}{l}{79.81\fontsize{7pt}{7pt}\selectfont±0.27} &
  58.02\fontsize{7pt}{7pt}\selectfont±1.12 &
  \multicolumn{1}{l}{90.39\fontsize{7pt}{7pt}\selectfont±0.15} &
  \multicolumn{1}{l}{88.40±\fontsize{7pt}{7pt}\selectfont0.07} &
  \multicolumn{1}{l}{85.31\fontsize{7pt}{7pt}\selectfont±0.17} &
  79.04\fontsize{7pt}{7pt}\selectfont±0.10 \\ \hline
\textbf{\ant~ } &
  {91.91\fontsize{7pt}{7pt}\selectfont±0.06} &
  \multicolumn{1}{l}{\textbf{90.34\fontsize{7pt}{7pt}\selectfont±0.16}} &
  \multicolumn{1}{l}{\textbf{87.71\fontsize{7pt}{7pt}\selectfont±0.17}} & %% cosine: 86.61±0.28
  \multicolumn{1}{l}{\textbf{83.17\fontsize{7pt}{7pt}\selectfont±0.53}} &
  \textbf{60.68\fontsize{7pt}{7pt}\selectfont±1.08}  & %% without bisection: 60.37±0.82
  \multicolumn{1}{l}{\textbf{91.40\fontsize{7pt}{7pt}\selectfont±0.14}} &
  \multicolumn{1}{l}{\textbf{89.90\fontsize{7pt}{7pt}\selectfont±0.06}} &
  \multicolumn{1}{l}{\textbf{87.95\fontsize{7pt}{7pt}\selectfont±0.28}} &
  \textbf{84.60\fontsize{7pt}{7pt}\selectfont±0.50} \\

% Antidote(alpha) &
%   ± &
%   \multicolumn{1}{l|}{±} &
%   \multicolumn{1}{l|}{±} &
%   \multicolumn{1}{l|}{±} &
%   ± &
%   \multicolumn{1}{l|}{±} &
%   \multicolumn{1}{l|}{±} &
%   \multicolumn{1}{l|}{±} &
%   ± \\ \hline
\end{tabular}
\begin{tabular}{ll|llll|llll}
\toprule
\toprule
CIFAR-100 &
  \multicolumn{1}{c}{Clean} & %%\multirow{2}{*}{Clean}
  \multicolumn{4}{c}{Symmetric Noise Rate} &
  \multicolumn{4}{c}{Asymmetric Noise Rate} \\ %%\cline{1-1} \cline{3-10} 
Loss  &
  \multicolumn{1}{c}{} &
  \multicolumn{1}{c}{0.2} &
  \multicolumn{1}{c}{0.4} &
  \multicolumn{1}{c}{0.6} &
  \multicolumn{1}{c}{0.8} &
  \multicolumn{1}{c}{0.1} &
  \multicolumn{1}{c}{0.2} &
  \multicolumn{1}{c}{0.3} &
  \multicolumn{1}{c}{0.4} \\ \hline
CE &
  70.79\fontsize{7pt}{7pt}\selectfont±0.58 &
  \multicolumn{1}{l}{56.21\fontsize{7pt}{7pt}\selectfont±2.04} &
  \multicolumn{1}{l}{39.31\fontsize{7pt}{7pt}\selectfont±0.74} &
  \multicolumn{1}{l}{22.38\fontsize{7pt}{7pt}\selectfont±0.74} &
  7.33\fontsize{7pt}{7pt}\selectfont±0.10 &
  \multicolumn{1}{l}{65.10\fontsize{7pt}{7pt}\selectfont±0.74} &
  \multicolumn{1}{l}{58.26\fontsize{7pt}{7pt}\selectfont±0.31} &
  \multicolumn{1}{l}{49.99\fontsize{7pt}{7pt}\selectfont±0.54} &
  41.15\fontsize{7pt}{7pt}\selectfont±1.04 \\ 
FL &
  70.58\fontsize{7pt}{7pt}\selectfont±0.34 &
  \multicolumn{1}{l}{56.32\fontsize{7pt}{7pt}\selectfont±1.43} &
  \multicolumn{1}{l}{40.83\fontsize{7pt}{7pt}\selectfont±0.52} &
  \multicolumn{1}{l}{22.44\fontsize{7pt}{7pt}\selectfont±0.54} &
  7.68\fontsize{7pt}{7pt}\selectfont±0.37 &
  \multicolumn{1}{l}{65.00\fontsize{7pt}{7pt}\selectfont±0.46} &
  \multicolumn{1}{l}{58.12\fontsize{7pt}{7pt}\selectfont±0.44} &
  \multicolumn{1}{l}{51.16\fontsize{7pt}{7pt}\selectfont±1.32} &
  41.46\fontsize{7pt}{7pt}\selectfont±0.38 \\ 
GCE &
  70.57\fontsize{7pt}{7pt}\selectfont±0.25 &
  \multicolumn{1}{l}{64.55\fontsize{7pt}{7pt}\selectfont±0.36} &
  \multicolumn{1}{l}{56.60\fontsize{7pt}{7pt}\selectfont±1.61} &
  \multicolumn{1}{l}{45.19\fontsize{7pt}{7pt}\selectfont±0.92} &
  19.85\fontsize{7pt}{7pt}\selectfont±0.88 &
  \multicolumn{1}{l}{63.94\fontsize{7pt}{7pt}\selectfont±2.08} &
  \multicolumn{1}{l}{60.89\fontsize{7pt}{7pt}\selectfont±0.06} &
  \multicolumn{1}{l}{53.36\fontsize{7pt}{7pt}\selectfont±1.58} &
  40.82\fontsize{7pt}{7pt}\selectfont±0.85 \\ 
NLNL &
  68.72\fontsize{7pt}{7pt}\selectfont±0.60 &
  \multicolumn{1}{l}{46.99\fontsize{7pt}{7pt}\selectfont±0.91} &
  \multicolumn{1}{l}{30.29\fontsize{7pt}{7pt}\selectfont±1.64} &
  \multicolumn{1}{l}{16.60\fontsize{7pt}{7pt}\selectfont±0.90} &
  11.01\fontsize{7pt}{7pt}\selectfont±2.48 &
  \multicolumn{1}{l}{59.55\fontsize{7pt}{7pt}\selectfont±1.22} &
  \multicolumn{1}{l}{50.19\fontsize{7pt}{7pt}\selectfont±0.56} &
  \multicolumn{1}{l}{42.81\fontsize{7pt}{7pt}\selectfont±1.13} &
  35.10\fontsize{7pt}{7pt}\selectfont±0.20 \\ 
SCE &
  70.41\fontsize{7pt}{7pt}\selectfont±0.20 &
  \multicolumn{1}{l}{55.23\fontsize{7pt}{7pt}\selectfont±0.76} &
  \multicolumn{1}{l}{40.23\fontsize{7pt}{7pt}\selectfont±0.29} &
  \multicolumn{1}{l}{21.44\fontsize{7pt}{7pt}\selectfont±0.52} &
  7.63\fontsize{7pt}{7pt}\selectfont±0.24 &
  \multicolumn{1}{l}{64.54\fontsize{7pt}{7pt}\selectfont±0.30} &
  \multicolumn{1}{l}{57.62\fontsize{7pt}{7pt}\selectfont±0.70} &
  \multicolumn{1}{l}{50.17\fontsize{7pt}{7pt}\selectfont±0.19} &
  41.01\fontsize{7pt}{7pt}\selectfont±0.74 \\ 
NCE+MAE &
  67.69\fontsize{7pt}{7pt}\selectfont±0.05 &
  \multicolumn{1}{l}{63.21\fontsize{7pt}{7pt}\selectfont±0.44} &
  \multicolumn{1}{l}{57.91\fontsize{7pt}{7pt}\selectfont±0.45} &
  \multicolumn{1}{l}{45.26\fontsize{7pt}{7pt}\selectfont±0.44} &
  23.72\fontsize{7pt}{7pt}\selectfont±0.99 &
  \multicolumn{1}{l}{65.70\fontsize{7pt}{7pt}\selectfont±1.04} &
  \multicolumn{1}{l}{62.87\fontsize{7pt}{7pt}\selectfont±0.42} &
  \multicolumn{1}{l}{55.82\fontsize{7pt}{7pt}\selectfont±0.19} &
  41.86\fontsize{7pt}{7pt}\selectfont±0.27 \\ 
NCE+RCE &
  67.89\fontsize{7pt}{7pt}\selectfont±0.47 &
  \multicolumn{1}{l}{64.60\fontsize{7pt}{7pt}\selectfont±0.92} &
  \multicolumn{1}{l}{58.64\fontsize{7pt}{7pt}\selectfont±0.19} &
  \multicolumn{1}{l}{45.25\fontsize{7pt}{7pt}\selectfont±0.50} &
  24.87\fontsize{7pt}{7pt}\selectfont±0.52 &
  \multicolumn{1}{l}{66.20\fontsize{7pt}{7pt}\selectfont±0.28} &
  \multicolumn{1}{l}{63.18\fontsize{7pt}{7pt}\selectfont±0.37} &
  \multicolumn{1}{l}{55.05\fontsize{7pt}{7pt}\selectfont±0.32} &
  41.21\fontsize{7pt}{7pt}\selectfont±0.66 \\ 
NFL+RCE &
  68.28\fontsize{7pt}{7pt}\selectfont±0.30 &
  \multicolumn{1}{l}{64.57\fontsize{7pt}{7pt}\selectfont±0.52} &
  \multicolumn{1}{l}{57.64\fontsize{7pt}{7pt}\selectfont±0.74} &
  \multicolumn{1}{l}{45.47\fontsize{7pt}{7pt}\selectfont±0.59} &
  24.35\fontsize{7pt}{7pt}\selectfont±0.32 &
  \multicolumn{1}{l}{66.18\fontsize{7pt}{7pt}\selectfont±0.38} &
  \multicolumn{1}{l}{63.63\fontsize{7pt}{7pt}\selectfont±0.30} &
  \multicolumn{1}{l}{55.33\fontsize{7pt}{7pt}\selectfont±0.25} &
  40.82\fontsize{7pt}{7pt}\selectfont±0.67 \\ 
NCE+AUL &
  69.55\fontsize{7pt}{7pt}\selectfont\fontsize{7pt}{7pt}\selectfont±0.40 &
  \multicolumn{1}{l}{65.12\fontsize{7pt}{7pt}\selectfont±0.36} &
  \multicolumn{1}{l}{55.86\fontsize{7pt}{7pt}\selectfont±0.20} &
  \multicolumn{1}{l}{37.88\fontsize{7pt}{7pt}\selectfont±0.32} &
  12.69\fontsize{7pt}{7pt}\selectfont±0.14 &
  \multicolumn{1}{l}{67.06\fontsize{7pt}{7pt}\selectfont±0.23} &
  \multicolumn{1}{l}{58.16\fontsize{7pt}{7pt}\selectfont±0.17} &
  \multicolumn{1}{l}{48.06\fontsize{7pt}{7pt}\selectfont±0.16} &
  38.30\fontsize{7pt}{7pt}\selectfont±0.12 \\ 
NCE+AGCE &
  68.78\fontsize{7pt}{7pt}\selectfont±0.24 &
  \multicolumn{1}{l}{65.30\fontsize{7pt}{7pt}\selectfont±0.46} &
  \multicolumn{1}{l}{59.95\fontsize{7pt}{7pt}\selectfont±0.15} &
  \multicolumn{1}{l}{47.63\fontsize{7pt}{7pt}\selectfont±0.94} &
  24.13\fontsize{7pt}{7pt}\selectfont±0.06 &
  \multicolumn{1}{l}{67.15\fontsize{7pt}{7pt}\selectfont±0.40} &
  \multicolumn{1}{l}{64.21\fontsize{7pt}{7pt}\selectfont±0.17} &
  \multicolumn{1}{l}{56.18\fontsize{7pt}{7pt}\selectfont±0.24} &
  44.15\fontsize{7pt}{7pt}\selectfont±0.08 \\ 
NCE+AEL &
  64.47\fontsize{7pt}{7pt}\selectfont±0.19 &
  \multicolumn{1}{l}{48.07\fontsize{7pt}{7pt}\selectfont±0.16} &
  \multicolumn{1}{l}{32.29\fontsize{7pt}{7pt}\selectfont±0.71} &
  \multicolumn{1}{l}{19.78\fontsize{7pt}{7pt}\selectfont±1.03} &
  10.50\fontsize{7pt}{7pt}\selectfont±0.51 &
  \multicolumn{1}{l}{58.20\fontsize{7pt}{7pt}\selectfont±0.37} &
  \multicolumn{1}{l}{50.19\fontsize{7pt}{7pt}\selectfont±0.61} &
  \multicolumn{1}{l}{43.82\fontsize{7pt}{7pt}\selectfont±0.32} &
  35.13\fontsize{7pt}{7pt}\selectfont±0.23 \\ 
LDR-KL &
  71.03\fontsize{7pt}{7pt}\selectfont±0.28 &
  \multicolumn{1}{l}{56.69\fontsize{7pt}{7pt}\selectfont±0.06} &
  \multicolumn{1}{l}{40.69\fontsize{7pt}{7pt}\selectfont±0.66} &
  \multicolumn{1}{l}{22.59\fontsize{7pt}{7pt}\selectfont±0.23} &
  7.49\fontsize{7pt}{7pt}\selectfont±0.33 &
  \multicolumn{1}{l}{65.93\fontsize{7pt}{7pt}\selectfont±0.01} &
  \multicolumn{1}{l}{58.47\fontsize{7pt}{7pt}\selectfont±0.04} &
  \multicolumn{1}{l}{50.92\fontsize{7pt}{7pt}\selectfont±0.15} &
  41.94\fontsize{7pt}{7pt}\selectfont±0.37 \\ 
CE+LC &
  71.80\fontsize{7pt}{7pt}\selectfont±0.34 &
  \multicolumn{1}{l}{56.26\fontsize{7pt}{7pt}\selectfont±0.09} &
  \multicolumn{1}{l}{37.36\fontsize{7pt}{7pt}\selectfont±0.49} &
  \multicolumn{1}{l}{17.46\fontsize{7pt}{7pt}\selectfont±0.62} &
  6.32\fontsize{7pt}{7pt}\selectfont±0.16 &
  \multicolumn{1}{l}{65.85\fontsize{7pt}{7pt}\selectfont±0.30} &
  \multicolumn{1}{l}{58.84\fontsize{7pt}{7pt}\selectfont±0.02} &
  \multicolumn{1}{l}{50.46\fontsize{7pt}{7pt}\selectfont±0.12} &
  40.97\fontsize{7pt}{7pt}\selectfont±0.39 \\ 
CE${}_\epsilon$+MAE &
  70.83\fontsize{7pt}{7pt}\selectfont±0.18 &
  \multicolumn{1}{l}{65.45\fontsize{7pt}{7pt}\selectfont±0.31} &
  \multicolumn{1}{l}{59.20\fontsize{7pt}{7pt}\selectfont±0.42} &
  \multicolumn{1}{l}{48.15\fontsize{7pt}{7pt}\selectfont±0.79} &
  26.30\fontsize{7pt}{7pt}\selectfont±0.46 &
  \multicolumn{1}{l}{67.58\fontsize{7pt}{7pt}\selectfont±0.04} &
  \multicolumn{1}{l}{64.52\fontsize{7pt}{7pt}\selectfont±0.18} &
  \multicolumn{1}{l}{58.47\fontsize{7pt}{7pt}\selectfont±0.12} &
  48.51\fontsize{7pt}{7pt}\selectfont±0.36 \\ 
FL${}_\epsilon$+MAE &
  70.58\fontsize{7pt}{7pt}\selectfont±0.68 &
  \multicolumn{1}{l}{65.45\fontsize{7pt}{7pt}\selectfont±1.39} &
  \multicolumn{1}{l}{59.58\fontsize{7pt}{7pt}\selectfont±0.80} &
  \multicolumn{1}{l}{48.09\fontsize{7pt}{7pt}\selectfont±0.35} &
  26.73\fontsize{7pt}{7pt}\selectfont±0.45 &
  \multicolumn{1}{l}{67.73\fontsize{7pt}{7pt}\selectfont±0.12} &
  \multicolumn{1}{l}{64.80\fontsize{7pt}{7pt}\selectfont±0.29} &
  \multicolumn{1}{l}{58.88\fontsize{7pt}{7pt}\selectfont±0.30} &
  48.10\fontsize{7pt}{7pt}\selectfont±0.23 \\ \hline
\textbf{\ant~} &
  71.17\fontsize{7pt}{7pt}\selectfont±1.19 &
  \multicolumn{1}{l}{\textbf{66.95\fontsize{7pt}{7pt}\selectfont±0.47}} &
  \multicolumn{1}{l}{\textbf{62.01\fontsize{7pt}{7pt}\selectfont±0.29}} &
  \multicolumn{1}{l}{\textbf{51.10\fontsize{7pt}{7pt}\selectfont±0.71}} &
  \textbf{27.22\fontsize{7pt}{7pt}\selectfont±0.65} &
  \multicolumn{1}{l}{\textbf{68.57\fontsize{7pt}{7pt}\selectfont±1.34}} &
  \multicolumn{1}{l}{\textbf{65.87\fontsize{7pt}{7pt}\selectfont±0.29}} &
  \multicolumn{1}{l}{\textbf{61.90\fontsize{7pt}{7pt}\selectfont±0.73}} &
  \textbf{57.89\fontsize{7pt}{7pt}\selectfont±1.14} \\ \hline
% Antidote(alpha) &
%   ± &
%   \multicolumn{1}{l}{±} &
%   \multicolumn{1}{l}{±} &
%   \multicolumn{1}{l}{±} &
%   ± &
%   \multicolumn{1}{l}{±} &
%   \multicolumn{1}{l}{±} &
%   \multicolumn{1}{l}{±} &
%   ± \\ \hline
\end{tabular}
\vspace{-.3cm}
\end{table*}

\subsection{Benchmark Evaluations on Symmetric and Asymmetric Noise}
It is common in the literature to evaluate the performance of a new loss against its counterparts on synthetic datasets with both asymmetric and symmetric noise, including CIFAR-10 and CIFAR-100. We followed \citep{wangepsilon_2024_neurips, zhou2021asymmetric_loss, ma2020normalized_normalized} to conduct similar experiments to compare our \ant's performance against the numbers reported in \cite{wangepsilon_2024_neurips}.  In particular, for asymmetric label noise on CIFAR-10, the label corruptions occur as follows: TRUCK $\to$ AUTOMOBILE, BIRD $\to$ AIRPLANE, DEER $\to$ HORSE, and CAT $\leftrightarrow$ DOG. For CIFAR-100, the 100 classes are grouped into 20 super-classes, each containing 5 sub-classes, and the labels are flipped within the same super-class into the next one. Table~\ref{tab:benchmark} compares \ant's performance with leading comparable methods in the field. As observed in Table~\ref{tab:benchmark}, \ant~ clearly outperforms all current losses, including the highest accuracy obtained by the  recent $\epsilon$-softmax method \cite{wangepsilon_2024_neurips} in CE${}_\epsilon$+MAE and FL${}_\epsilon$+MAE, across all noise levels and for both CIFAR-10 and CIFAR-100. Another key observation is that for high noise ratios, the performance gap between \ant~ and its leading counterparts becomes even more significant. For example, on CIFAR-10 with 60\% and 80\% symmetric noise \ant~ achieves a 2.81\% and 1.72\% performance improvement over the leading competitors  NCE+AGCE  and CE${}_\epsilon$+MAE (e.g., 83.17\% from \ant~ vs. 80.36\% from NCE+AGCE at 60\% symmetric noise). For asymmetric noise, the performance gap is even more significant. For instance, there is a 5.17\% performance improvement over the highest performing competing method (84.60\% from \ant~ vs. 79.43\% from NCE+RCE) at 40\% asymmetric noise. A similar pattern is also observed for CIFAR-100. For example, at 40\% asymmetric noise \ant~ outperforms CE${}_\epsilon$+MAE by 9.38\% (57.89\% from \ant~ vs. 48.51\% from CE${}_\epsilon$+MAE). See Appendix \ref{app:transition_matrix_comp} for additional comparisons with transition-matrix based methods, where we also find that \ant~ outperforms on CIFAR-10 and CIFAR-100.  
\newpage
 
\subsection{Benchmark Evaluations on Large Real-World Noisy Datasets: WebVision and ImageNet}

Consistent with \citep{wangepsilon_2024_neurips, zhou2021asymmetric_loss, ma2020normalized_normalized}, we tested \ant~ on real-world large datasets collected from the web: WebVision and ImageNet. 
Table~\ref{tab:real} shows the evaluation results. There we see that \ant~ outperforms all the benchmark methods by a high margin in terms of top-1 and top-5 accuracies on both WebVision and ImageNet validation datasets. For instance, \ant~ outperforms CE${}_\epsilon$+MAE by 1.12\% in top-1 accuracy and 1.68\% in top-5 accuracy on ImageNet. These results suggest that the principled information divergence-based forgetting mechanism in \ant~ is an  effective tool in combating label noise in both synthetic and real world datasets and across all noise levels.

\begin{table*}[h!]
\small
\centering
\setlength\tabcolsep{5.5 pt}
\caption{Last epoch accuracies (\%) on the WebVision and ImageNet (ILSVRC12) validation sets. Best performance appear in bold font.}
\label{tab:real}

\begin{tabular}{cllllllllll}
\toprule
\multicolumn{2}{c}{Method}         & \multicolumn{1}{c}{CE}    & \multicolumn{1}{c}{GCE}   & \multicolumn{1}{c}{SCE}   & \multicolumn{1}{c}{AGCE}  & \multicolumn{1}{c}{\shortstack{NCE+\\RCE}} & \multicolumn{1}{c}{\shortstack{NCE+\\AGCE}} & \multicolumn{1}{c}{\shortstack{LDR\\-KL}} & \multicolumn{1}{c}{\shortstack{CE${}_\epsilon$+\\MAE}} & \multicolumn{1}{c}{\textbf{\ant}} \\
\toprule
\multirow{2}{*}{WebVision} & Top-1 & 66.08 & 61.96 & 67.92 & 69.48 & 66.88   & 66.00    & 69.64  & 71.32       & \textbf{72.12\fontsize{7pt}{7pt}\selectfont±0.69}  \\
                           & Top-5 & 84.76 & 76.80 & 86.36 & 87.28 & 86.48   & 85.20    & 87.16  & 88.48       & \textbf{89.04\fontsize{7pt}{7pt}\selectfont±0.48}  \\

\toprule
\multicolumn{1}{l}{\multirow{2}{*}{ImageNet}} & Top-1 & 60.72 & 60.52 & 63.28 & 65.12 & 63.96 & 62.68 & 65.24 & 67.20 & \textbf{68.32\fontsize{7pt}{7pt}\selectfont±0.65} \\
\multicolumn{1}{l}{}       & Top-5 & 84.76 & 76.56 & 85.16 & 86.12 & 84.68   & 84.96    & 86.12  & 87.48       & \textbf{89.16\fontsize{7pt}{7pt}\selectfont±0.58} \\ \hline
\end{tabular}
\vspace{-.2cm}
\end{table*}

\subsection{Benchmark Evaluations on the Human-Annotated CIFAR-100N}
Following \cite{wangepsilon_2024_neurips}, we also evaluate \ant~ on human-annotated datasets CIFAR-100N, which is often considered a more difficult noise than symmetric or asymmetric noise. Consistent with \cite{wangepsilon_2024_neurips}, we compare \ant~ against standard CE, GCE\cite{zhang2018generalized_GCE_loss}, Forward T \cite{patrini2017making_ForwardT}, F-Div \cite{wei2021optimizing_F-Div}, Peer Loss \cite{liu2020peer_PeerLoss}, VolMinNet \cite{li2021provablyVolMinNet}, Negative-LS \cite{wei2022smooth_Negative_LS}, and CE${}_\epsilon$+MAE \cite{wangepsilon_2024_neurips}. Table \ref{tab:natural} summarizes the results of our comparison for 5 random runs. Again we find that \ant~ outperforms the competing methods.

\begin{table}[!h]
\vspace{-.3cm}
\small
\centering
\setlength\tabcolsep{1.1 pt}
\caption{Best epoch test accuracies (\%) on the CIFAR-100N dataset. Following the settings in \cite{wangepsilon_2024_neurips}, the results "mean±std"
are reported over 5 random runs and the best results are boldfaced.}
\label{tab:natural}
\begin{tabular}{llllllllll}
\toprule
Method                                                  & \multicolumn{1}{c}{CE}     & \multicolumn{1}{c}{GCE}        & \multicolumn{1}{c}{Forward T}  & \multicolumn{1}{c}{F-Div}      & \multicolumn{1}{c}{Peer Loss}  & \multicolumn{1}{c}{VolMinNet}  & \multicolumn{1}{c}{\shortstack{Negative-\\LS}} & \multicolumn{1}{c}{\shortstack{CE${}_\epsilon$\\+MAE}}     & \multicolumn{1}{c}{\textbf{\ant}}        \\ \hline
\begin{tabular}[c]{@{}l@{}}Acc. (\%)\end{tabular} & {55.50\fontsize{7pt}{7pt}\selectfont±0.66} & {56.73\fontsize{7pt}{7pt}\selectfont±0.30} & {57.01\fontsize{7pt}{7pt}\selectfont±1.03} & {57.10\fontsize{7pt}{7pt}\selectfont±0.65} & {57.59\fontsize{7pt}{7pt}\selectfont±0.61} & {57.80\fontsize{7pt}{7pt}\selectfont±0.31} & {58.59\fontsize{7pt}{7pt}\selectfont±0.98}  & {61.78\fontsize{7pt}{7pt}\selectfont±0.14} & \textbf{62.17\fontsize{7pt}{7pt}\selectfont±0.23} \\ \hline
\end{tabular}
\vspace{-.4cm}
\end{table}

\subsection{Benchmark Evaluations with Add-ons}\label{sec:add_ons}
Following \cite{wangepsilon_2024_neurips}, we also evaluate \ant~ on human-annotated datasets CIFAR-100N, when add-on techniques such as semi-supervised learning, dual-network training, and MixUp are employed for all methods to ensure fair comparisons.  Consistent with \cite{wangepsilon_2024_neurips}, we compare \ant~ against ELR+\cite{liu2020early_ELR}, SOP+ \cite{liu2022robust_SOP}, Divide-Mix \cite{Li2020DivideMix}, Proto-semi \cite{zhu2023rethinking_ProtoSemi}, PES (Semi) \cite{bai2021understanding_PES}, CORES* \cite{cheng2021learning_CORES}, CAL \cite{zhu2021second_CAL}, and CE${}_\epsilon$+MAE \cite{wangepsilon_2024_neurips}. Table \ref{tab:add_ons} summarizes the results of our comparison for 5 random runs. Again, we find that \ant~ outperforms the competing methods.

%% Coteaching+ \cite{yu2019does_COTeachingPlus}

\begin{table}[!h]
\vspace{-.3cm}
\small
\centering
\setlength\tabcolsep{1.1 pt}
\caption{Best epoch test accuracies (\%) on the CIFAR-100N dataset. Following the settings in \cite{wangepsilon_2024_neurips}, the results "mean±std"
are reported over 5 random runs and the best results are boldfaced.}
\label{tab:add_ons}
\begin{tabular}{llllllllll}
\toprule
Method                                                  & \multicolumn{1}{c}{CAL}         & \multicolumn{1}{c}{CORES*}        & \multicolumn{1}{c}{DivideMix}  & \multicolumn{1}{c}{ELR+}      & \multicolumn{1}{c}{\shortstack{PES\\(Semi)}}  & \multicolumn{1}{c}{Proto-smi}  & \multicolumn{1}{c}{SOP+} & \multicolumn{1}{c}{\shortstack{CE${}_\epsilon$\\+MAE}}     & \multicolumn{1}{c}{\textbf{\ant}}        \\ \hline
\begin{tabular}[c]{@{}l@{}}Acc. (\%)\end{tabular} & {61.73\fontsize{7pt}{7pt}\selectfont±0.42} & {55.72\fontsize{7pt}{7pt}\selectfont±0.42} & {71.13\fontsize{7pt}{7pt}\selectfont±0.48} & {66.72\fontsize{7pt}{7pt}\selectfont±0.07} & {70.36\fontsize{7pt}{7pt}\selectfont±0.33} & {67.73\fontsize{7pt}{7pt}\selectfont±0.67} & {67.81\fontsize{7pt}{7pt}\selectfont±0.23}  & {71.97\fontsize{7pt}{7pt}\selectfont±0.18} & \textbf{72.33\fontsize{7pt}{7pt}\selectfont±0.28} \\ \hline
\end{tabular}
\vspace{-.4cm}
\end{table}

\subsection{Empirical Time Comparison of \ant}
\label{app:time}

%% Markdown Format for OpenReview
% **Table: Best epoch test accuracies (\%) on the CIFAR-100N dataset. Following the settings in (Wang et al. 2024), the results \"mean±std\" are reported over 5 random runs and the best results are boldfaced.**

% | Method            | CAL (\%)        | CORES* (\%)     | DivideMix (\%)  | ELR+ (\%)        | PES(Semi) (\%)   | Proto-smi (\%)   | SOP+ (\%)        | CE$_\epsilon$+MAE (\%) | **ANT** (\%)        |
% |-------------------|------------------|------------------|------------------|------------------|------------------|------------------|------------------|---------------------------|---------------------|
% | Accuracy (\%)     | 61.73 ± 0.42     | 55.72 ± 0.42     | 71.13 ± 0.48     | 66.72 ± 0.07     | 70.36 ± 0.33     | 67.73 ± 0.67     | 67.81 ± 0.23     | 71.97 ± 0.18              | **72.33 ± 0.28**    |

Table \ref{tab:time} summarizes the empirical time complexity comparison of the proposed \ant~ method  with the classic non-robust cross-entropy (CE) loss, as well as the recent counterpart $\epsilon$-softmax, on an RTX 4090 GPU with 24 GB memory. As shown in the table, the elapsed training time of \ant~ is very close to that of CE (less than 25 seconds difference on both CIFAR-10 and CIFAR-100). This efficiency is mainly due to the fact that the inner optimization in \ant~ is a convex problem in (at most) 2 variables ($\lambda$ and $\rho$ in general, only $\lambda$  for the KL variant), and thus adds little to the computational cost. Furthermore, we find that \ant~ has a better runtime than  $\epsilon$-softmax.

\begin{table}[!h]
\vspace{-.3cm}
\small
\centering
\caption{Empirical time comparison of \ant~ and its counterparts (minutes:seconds) }
\small
\label{tab:time}
\begin{tabular}{llll}\toprule
Method                &  CE (non-robust)   &  $\epsilon$-softmax &  \ant  \\\hline
CIFAR-10 (120 epochs) & 3:36    &4:12 &3:52   \\ 
CIFAR-100 (200 epochs) & 33:28    &34:14 & 33:52   \\ 
\hline              
\end{tabular}
\vspace{-.4cm}
\end{table}

\section{Conclusion}
We proposed a new  approach to deep learning with noisy labels that uses relaxation-optimization over information divergence neighborhoods. Based on this new formulation, we introduced \ant, a computationally efficient solution formulated in terms of a new class of $f$-divergence-based loss functions along with a low-dimensional adversarial convex optimization problem. These new losses enjoy a time complexity that is very close to that of the standard cross entropy loss. Through extensive evaluations, we showed that \ant~ yields leading performance in the field. We evaluated \ant~ on five datasets,  CIFAR-10, CIFAR-100, CIFAR-100N, WebVision, and ImageNet with symmetric, asymmetric synthetic, human annotation, and real-world label noise. As most real-world datasets are expected to be noisy, our results suggest that \ant~ can be an important ingredient in the development of robust and reliable AI systems.   A limitation of the proposed method  is the need to select the hyperparameter $\delta$ (or alternatively, the more easily tuned  penalty strength $C$ in the approach outlined in Appendix \ref{app:delta_penalized_opt}). Moreover, while our general theory applies to general loss functions, we restricted attention to the CE loss in our experiments.  Exploring the performance of \ant~ when combined with other loss functions is an interesting direction for further exploration.

%\begin{ack}

%\end{ack}

\bibliography{antidote_arxiv.bbl}

%%%%%%%%%%%%%%%%%%%%%%%%%%%%%%%%%%%%%%%%%%%%%%%%%%%%%%%%%%%%

\appendix

\section{Proofs}

\subsection{Relaxed Problem Solution Set}\label{app:thm_optimizer}
Here we provide a detailed proof of Theorem \ref{thm:exact_sol_main}, which relates the solution to the relaxed problem \eqref{eq:relaxed_optim} in the presence of label noise to the solution of standard training \eqref{eq:stoch_opt} when there is no label noise. Rather than working with empirical distributions, for these theoretical results we work with the exact and noisy distributions defined as follows.
\begin{definition}\label{def:P_star_P_T}
Given a probability distribution $P_\mathcal{X}$ on the sample space $\mathcal{X}$  with deterministic labels defined by the measurable map $h_*:\mathcal{X}\to \mathcal{Y}$ ( where $\mathcal{Y}$ is a finite set of labels), we define $P_{*}$ to be the corresponding distribution on  $\mathcal{X}\times\mathcal{Y}$, the space of sample/label pairs, i.e.,
\begin{align}\label{eq:P_star_def}
P_{*}\coloneqq\delta_{h_*(x)}(dy) P_{\mathcal{X}}(dx)\,.
\end{align}

Given a measurable  sample-dependent transition matrix, $T^x_{y,\tilde{y}}$ (i.e., the probability of the sample $x$ having its label, $y$,  flipped to the corrupted label $\tilde{y}$), we  define the corresponding distribution on samples and corrupted labels  by
\begin{align}\label{eq:P_T_def}
 P_{T}\coloneqq \sum_{\tilde{y}\in\mathcal{Y}} T^x_{h_*(x),\tilde y}\delta_{\tilde{y}}(dy) P_{\mathcal{X}}(d{x})\,.  
\end{align}
\end{definition}
First we prove that, under mild assumptions, the exact labeling map $h_*$ provides a solution to the relaxed problem \eqref{eq:relaxed_optim}, with  $P_n$ replaced by $P_T$.
\begin{theorem}\label{thm:exact_labels_solve_relaxed_problem}
Suppose that $P_{\mathcal{X}}$, $h_*$, $P_*$, and $P_T$ are as in Definition \ref{def:P_star_P_T} and also:
\begin{enumerate}
    \item  $\inf_{x\in\mathcal{X}}T_{h_*(x),h_*(x)}^x>0$\,.
\item We have a classifier family $h_\theta:\mathcal{X}\to\mathcal{P}(\mathcal{Y})$, $\theta\in \Theta$,  where $\mathcal{P}(\mathcal{Y})$ denotes the space of probability vectors over the class labels. We also suppose that there exists  $\theta_*\in\Theta$ such that $h_{\theta_*}=h_*$ (more precisely, this means $h_{\theta_*}=1_{h_*}$, but to simplify notation we will identify each label with its corresponding one-hot vector).
\item We have a loss function ${L}:\mathcal{P}(\mathcal{Y})\times \mathcal{Y}\to [0,\infty]$ such that ${L}(p,y)=0$ if and only if $p=y$.  The training objective loss is defined by $\mathcal{L}_\theta(x,y)\coloneqq L(h_\theta(x),y)$.
\item We have a convex function  $f:(0,\infty)\to\mathbb{R}$ such that $f(1)=0$ and $f(0)\coloneqq \lim_{t\to0^+}f(t)$ is finite.
\end{enumerate}

Then there exists $\delta\in(0,\infty)$ such that
\begin{align}\label{eq:equality_of_perturbed_and_unperturbed}
\inf_{\theta\in\Theta} \inf_{Q: D_f(Q\|P_T)\leq \delta} E_{Q}[\mathcal{L}_\theta]=0\,,
\end{align}
with  $\theta_*$ being a minimizer, i.e., the relaxed optimization problem is solved by the exact non-noisy classifier $h_{\theta_*}=h_*$. 
\end{theorem}
\begin{remark}
    Note that when applying the definition \ref{eq:reformulation_final_KL}, the function $f$ should be continuously extended to $0$, as per the above definition of $f(0)$. For instance, $f_{\mathrm{KL}}(0)=0$.
\end{remark}
\begin{proof}
Our assumptions imply $T^x_{h_*(x),h_*(x)}>0$ and so,  from the definitions \eqref{eq:P_star_def} and \eqref{eq:P_T_def},  we see that $P_*\ll P_T$ with 
\begin{align}
 \frac{dP_*}{dP_T}=\frac{1_{h_*(x)}(y)}{T_{h_*(x),h_*(x)}^x}\,.
\end{align}
Using the definition \eqref{eq:f_div_def} we can then compute
\begin{align}
    D_f(P_*\|P_T)=&\int f\left(\frac{1_{h_*(x)}(y)}{T_{h_*(x),h_*(x)}^x}\right) P_T(dxdy)
    =\int \sum_{\tilde{y}\in\mathcal{Y}} T^x_{h_*(x),\tilde{y}}f\left(\frac{1_{h_*(x)}(\tilde{y})}{T_{h_*(x),h_*(x)}^x}\right) P_{\mathcal{X}}(dx)\\
    =&\int \left(1- T^x_{h_*(x),h_*(x)}\right)f\left(0\right) +T^x_{h_*(x),h_*(x)}f\left(1/T_{h_*(x),h_*(x)}^x\right) P_{\mathcal{X}}(dx)\notag\,.
\end{align}
Recalling the assumptions that $f(0)$ is finite and $q\coloneqq \inf_{x\in\mathcal{X}}T_{h_*(x),h_*(x)}^x>0$, we therefore have
\begin{align}
D_f(P_*\|P_T)\leq   \max\{f(0), \sup_{t\in[1,1/q]}f(t)\}<\infty\,, 
\end{align}
where finiteness of the bound from the fact that $f$ is continuous on $(0,\infty)$, due to it being convex.

The above computation implies that there exists $\delta\in(0,\infty)$ such that $D_f(P_*\|P_T)\leq  \delta$  and for such $\delta$ we have 
\begin{align}
&\inf_{\theta\in\Theta} \inf_{Q: D_f(Q\|P_T)\leq \delta} E_{Q}[\mathcal{L}_\theta]\leq E_{P_*}[\mathcal{L}_{\theta_*}]=\int L(h_{\theta_*}(x),y) P_*(dxdy)\\
=&\int L(h_{\theta_*}(x),h_*(x)) P_{\mathcal{X}}(dx)=\int L(h_{*}(x),h_*(x)) P_{\mathcal{X}}(dx)=0\,.\notag
\end{align}
Together with non-negativity of $\mathcal{L}_\theta$, this implies  \eqref{eq:equality_of_perturbed_and_unperturbed} and also that $\theta_*$ is a minimizer.
\end{proof}

With additional assumptions, we can prove Theorem \ref{thm:exact_sol_main} from the main text, i.e., we prove that the exact labeling map $h_*$ is the unique solution to the relaxed problem.
\begin{theorem}\label{thm:h_star_unique_sol_app}
Let $\mathcal{X}$ be a complete separable metric space with its Borel sigma algebra and let $P_{\mathcal{X}}$, $h_*$, $P_*$,  $P_T$ be as in Definition \ref{def:P_star_P_T}. Suppose the following.
\begin{enumerate}
\item We have $r\in(0,1)$ such that  $T^x_{h^*(x),h^*(x)}=1-r$ for all $x\in\mathcal{X}$ and
\begin{align}\label{eq:sup_T_r_bound}
   T^x_{h_*(x),y}<1-r\, \text{ for all $x\in\mathcal{X},y\in\mathcal{Y}$  that satisfy $y\neq h_*(x)$}\,.
\end{align}
Note that the latter condition automatically follows from the former if $r<1/2$.
\item We have a classifier family $h_\theta:\mathcal{X}\to\mathcal{P}(\mathcal{Y})$, $\theta\in \Theta$,  where $\mathcal{P}(\mathcal{Y})$ denotes the space of probability vectors over the class labels. We also suppose that there exists $\theta_*\in\Theta$ such that $h_{\theta_*}=h_*$ (again, note that  we identify each label with its corresponding one-hot vector).
\item We have a loss function ${L}:\mathcal{P}(\mathcal{Y})\times \mathcal{Y}\to [0,\infty]$ such that ${L}(p,y)=0$ if and only if $p=y$. 
\item Equip $\mathcal{Y}$ with the discrete metric, $d_{\mathcal{Y}}(y_1,y_2)\coloneqq 1_{y_1\neq y_2}$ and assume that $\mathcal{L}_\theta(x,y)\coloneqq L(h_\theta(x),y)$ is lower semicontinuous (LSC) on $\mathcal{X}\times\mathcal{Y}$ for all $\theta\in\Theta$.
\item We have  a strictly convex $f:(0,\infty)\to \mathbb{R}$ such that  $f(1)=0$, $f(0)\coloneqq \lim_{t\to 0^+} f(t)$ is finite, and  $f^*(y)$ is finite for all $y\in\mathbb{R}$.
\end{enumerate}
Let $\delta\coloneqq rf(0)+(1-r)f(1/(1-r))$.  Then for all
\begin{align}\label{eq:argmin_relaxed}
 \tilde{\theta} \in\argmin_{ \theta\in\Theta}\inf_{Q: D_f(Q\|P_T)\leq \delta} E_{Q}[\mathcal{L}_\theta] 
\end{align}
  we have $h_{\tilde{\theta}}=h_*$ almost surely under $P_{\mathcal{X}}$ (abbreviated as $P_{\mathcal{X}}$-a.s.).
\end{theorem}
\begin{remark}
    Note that   item 5 of the above assumptions holds for $f_{\mathrm{KL}}$ and for $f_\alpha$ for $\alpha>1$.
\end{remark}
\begin{proof}
From the computations in the proof of Theorem \ref{thm:exact_labels_solve_relaxed_problem} we obtain
\begin{align}
    D_f(P_*\|P_T)=rf(0)+(1-r)f(1/(1-r))
\end{align}
Note that strict convexity of $f$ implies
\begin{align}
   \delta\coloneqq rf(0)+(1-r)f(1/(1-r))>f(1)=0\,, 
\end{align}
and so $\delta\in(0,\infty)$.
 Therefore the proof of Theorem \ref{thm:exact_labels_solve_relaxed_problem} implies
\begin{align}\label{eq:inf_Df_zero}
&\inf_{\theta\in\Theta} \inf_{Q: D_f(Q\|P_T)\leq \delta} E_{Q}[\mathcal{L}_\theta]=0\,.
\end{align}
  Now  fix  $\tilde{\theta}$ satisfying \eqref{eq:argmin_relaxed}. The equality \eqref{eq:inf_Df_zero}  implies
\begin{align}\label{eq:inf_Df_zero2}  
 \inf_{Q: D_f(Q\|P_T)\leq \delta} E_{Q}[\mathcal{L}_{\tilde\theta}]=0   \,.
\end{align}
In \eqref{eq:inf_Df_zero2}, $Q$ and $P_T$ are probability distributions on the  complete separable metric space
$\mathcal{X}\times \mathcal{Y}$ (with its Borel sigma algebra). Together with the finiteness of $f^*$, this implies $Q\mapsto D_f(Q\|P_T)$ is LSC and $\{Q: D_f(Q\|P_T)\leq \delta\}$ is compact (topological statements refer to the topology of weak convergence); see, e.g., Corollary 51  and Lemma 54 in \cite{birrell2022f}.  Therefore there exists $Q_j,\widetilde{Q}\in\{Q:D_f(Q\|P_T)\leq \delta\}$ such that $Q_j\to \widetilde{Q}$ weakly and $E_{Q_j}[\mathcal{L}_{\tilde{\theta}}]\to 0$.  By assumption, $\mathcal{L}_{\tilde{\theta}}$ is LSC and bounded below, hence the Portmanteau theorem implies $\lim_{j\to\infty} E_{Q_j}[ \mathcal{L}_{\tilde\theta}]\geq E_{\widetilde{Q}}[\mathcal{L}_{\tilde\theta}]$. From this we can conclude that there exists $\widetilde{Q}$ such that  $D_f(\widetilde{Q}\|P_T)\leq \delta$ and
\begin{align}\label{eq:E_Q_tilde_L_zero}
    E_{\widetilde{Q}}[\mathcal{L}_{\tilde\theta}]=0\,.
\end{align} 
Combining \eqref{eq:E_Q_tilde_L_zero} with the assumptions on the loss function (specifically,  non-negativity and the form of its zero-set) we find that
\begin{align}
\widetilde{Q}(\{(x,y):h_{\tilde{\theta}}(x)\neq y\})=0\,,
\end{align}
and so there exists  a measurable $\tilde{h}:\mathcal{X}\to \mathcal{Y}$ such that $\tilde{h}=h_{\tilde{\theta}}$ \, $\widetilde{Q}_\mathcal{X}$-a.s. and
\begin{align}\label{eq:Q_tilde_form}
    \widetilde{Q}(dxdy)=\delta_{\tilde{h}(x)}(dy) \widetilde{Q}_{\mathcal{X}}(dx)\,,
\end{align}
where $\widetilde{Q}_{\mathcal{X}}$ denotes the $\mathcal{X}$-marginal distribution of $\widetilde{Q}$.  The fact that $\widetilde{Q}\ll P_T$ allows us to compute
\begin{align}\label{eq:Q_tilde_T_hstar_htilde_0}
    \widetilde{Q}_{\mathcal{X}}(\{x:T^x_{h_*(x),\tilde{h}(x)}=0\})=&\int 1_{T^x_{h_*(x),y}=0}\widetilde{Q}(dxdy)\\
    =&\int 1_{T^x_{h^*(x),y}=0}\frac{d\widetilde{Q}}{dP_T}(x,y) P_T(dxdy)\notag\\
    =&\int \sum_{\tilde{y}} T^x_{h_*(x),\tilde{y}}1_{T^x_{h^*(x),\tilde{y}}=0}\frac{d\widetilde{Q}}{dP_T}(x,\tilde{y}) P_{\mathcal{X}}(dx)=0\notag\,,
\end{align}
and hence also
\begin{align}\label{eq:P_T_dQdP}
P_{\mathcal{X}}\left(\left\{x:T^x_{h^*(x),\tilde{h}(x)}=0,\frac{d\widetilde{Q}_{\mathcal{X}}}{dP_{\mathcal{X}}}(x)\neq 0\right\}\right)=0\,.
\end{align}
Therefore we have  
\begin{align}
    \frac{d\widetilde{Q}}{dP_T}(x,y)= \frac{1_{\tilde{h}(x)}(y)}{T^x_{h_*(x),\tilde{h}(x)}}\frac{d\widetilde{Q}_{\mathcal{X}}}{dP_{\mathcal{X}}}(x)
\end{align}
and hence we can  compute
\begin{align}\label{eq:Df_delta_bound}
    \delta\geq &D_f(\widetilde{Q}\|P_T)
    =\int f\left(\frac{d\widetilde{Q}}{dP_T}\right) dP_T\\
    =&\int  (1- T^x_{h_*(x),\tilde{h}(x)}) f\left(0\right)
    +T^x_{h_*(x),\tilde{h}(x)} f\left(\frac{1}{T^x_{h_*(x),\tilde{h}(x)}}\frac{d\widetilde{Q}_{\mathcal{X}}}{dP_{\mathcal{X}}}(x)\right) P_{\mathcal{X}}(dx)\notag\\
    =&\int_{h_*=\tilde{h}}  r f\left(0\right)
    +(1-r) f\left((1-r)^{-1}\frac{d\widetilde{Q}_{\mathcal{X}}}{dP_{\mathcal{X}}}(x)\right) P_{\mathcal{X}}(dx)\notag\\
    &+\int_{h_*\neq \tilde{h}}  f\left(0\right) +T^x_{h_*(x),\tilde{h}(x)} \left(f\left(\frac{1}{T^x_{h_*(x),\tilde{h}(x)}}\frac{d\widetilde{Q}_{\mathcal{X}}}{dP_{\mathcal{X}}}(x)\right) -  f\left(0\right)\right)
    P_{\mathcal{X}}(dx)\,.\notag
\end{align}
Convexity of $f$ implies that $(f(tc)-f(0))/t\geq (f(sc)-f(0))/s$ for all $0<s\leq t$ and all $c\geq 0$.  Applying this to \eqref{eq:Df_delta_bound}, noting that $T^x_{h_*(x),\tilde{h}(x)}<1-r$ on $\{h_*\neq \tilde{h}\}$, and using \eqref{eq:P_T_dQdP} we therefore obtain
\begin{align}\label{eq:Df_delta_bound2}
    \delta\geq &D_f(\widetilde{Q}\|P_T)\\
    \geq &\int_{h_*=\tilde{h}}  r f\left(0\right)
    +(1-r) f\left((1-r)^{-1}\frac{d\widetilde{Q}_{\mathcal{X}}}{dP_{\mathcal{X}}}(x)\right) P_{\mathcal{X}}(dx)\notag\\
    &+\int_{h_*\neq \tilde{h}}  f\left(0\right) +1_{T_{h^*(x),\tilde{h}(x)}^x\neq 0} (1-r) \left(f\left((1-r)^{-1}\frac{d\widetilde{Q}_\mathcal{X}}{dP_{\mathcal{X}}}(x)\right)-f(0)\right)
    P_{\mathcal{X}}(dx)\notag\\
    =&\int r f\left(0\right)
    +(1-r) f\left((1-r)^{-1}\frac{d\widetilde{Q}_{\mathcal{X}}}{dP_{\mathcal{X}}}(x)\right) P_{\mathcal{X}}(dx)\notag\\
    =&\delta+\int 
    (1-r) \left(f\left((1-r)^{-1}\frac{d\widetilde{Q}_{\mathcal{X}}}{dP_{\mathcal{X}}}(x)\right)-f(1/(1-r))\right) P_{\mathcal{X}}(dx)\,.\notag
\end{align}
Define $f_r(t)\coloneqq(1-r)(f(t/(1-r))-f(1/(1-r)))$ (not to be mistaken for the function defining the $\alpha$ divergences).  Then $f_r$ is strictly convex and $f_r(1)=0$, hence the divergence property gives
\begin{align}
    \int 
    (1-r) \left(f\left((1-r)^{-1}\frac{d\widetilde{Q}_{\mathcal{X}}}{dP_{\mathcal{X}}}(x)\right)-f(1/(1-r))\right) P_{\mathcal{X}}(dx)
    =D_{f_r}(\widetilde{Q}_{\mathcal{X}}\|P_{\mathcal{X}})\geq 0
\end{align}
with equality if and only if $\widetilde{Q}_{\mathcal{X}}=P_{\mathcal{X}}$.  Combined with the bound \eqref{eq:Df_delta_bound2} we can therefore conclude $\widetilde{Q}_{\mathcal{X}}=P_{\mathcal{X}}$ and hence from \eqref{eq:Q_tilde_T_hstar_htilde_0} that $T^x_{h_*(x),\tilde{h}(x)}>0$ \, $P_{\mathcal{X}}$-a.s.  Now,  calculating similarly to \eqref{eq:Df_delta_bound}-\eqref{eq:Df_delta_bound2} but taking these facts into account, we obtain
\begin{align}\label{eq:Df_delta_bound3}
  &\int_E  \left(T^x_{h_*(x),\tilde{h}(x)} \left(f\left(\frac{1}{T^x_{h_*(x),\tilde{h}(x)}}\right) -  f\left(0\right)\right)
   - (1-r)(f(1/(1-r))-f\left(0\right)
    ) \right)P_{\mathcal{X}}(dx)\leq 0\,,\\
    &E\coloneqq\{x:h_*(x)\neq \tilde{h}(x),T^x_{h_*(x),\tilde{h}(x)}>0\}\,.\notag
\end{align}
Using the fact that strict convexity of $f$ implies $(f(s)-f(0))/s<(f(t)-f(0))/t$ for all $0<s<t$ along with the bound $T^x_{h_*(x),\tilde{h}(x)}<1-r$ when $h_*(x)\neq \tilde{h}(x)$  we see that the integrand in \eqref{eq:Df_delta_bound3} is strictly positive on $E$ and therefore we must have $P_{\mathcal{X}}(E)=0$.  As $T^x_{h_*(x),\tilde{h}(x)}>0$ \, $P_{\mathcal{X}}$-a.s., we  conclude that $h_*=\tilde{h}$\,$P_{\mathcal{X}}$-a.s.  We already showed that $\tilde{h}=h_{\tilde{\theta}}$ \, $\widetilde{Q}_\mathcal{X}$-a.s. and that $\widetilde{Q}_\mathcal{X}=P_{\mathcal{X}}$, thus we obtain $h_{\tilde{\theta}}=h_*$  $P_{\mathcal{X}}$-a.s. as claimed.
\end{proof}

\subsection{Reformulation via Convex Duality}\label{app:duality_proof}
For completeness, in this appendix we provide a detailed proof of the convex duality result, Theorem \ref{thm:duality}, which a special case of the following theorem.
\begin{theorem}
Let $\mathcal{Z}$ be a measurable space $P_n$ be the empirical distribution of the samples $z_i\in\mathcal{Z}$, $i=1,...,n$, and suppose we have:
\begin{enumerate}
    \item $0\leq a<1<b\leq \infty$ and a convex $f:(a,b)\to\mathbb{R}$ that satisfies $f(1)=0$.
    \item A measurable function $\mathcal{L}_\theta:\mathcal{Z}\to\mathbb{R}$, depending on parameters $\theta\in\Theta$.
\end{enumerate}
Then for all $\delta>0$, $\kappa\geq 0$ we have 
    \begin{align}\label{eq:reformulation_final_app}
&\inf_{\theta\in\Theta}\inf_{Q: D_f(Q\|P_n)\leq \delta}\{E_Q[\mathcal{L}_\theta]+\kappa D_f(Q\|P_n)\}\\
=&\inf_{\theta\in\Theta}\sup_{\substack{\lambda>0,\\\rho\in\mathbb{R}} }\!\left\{\!-\lambda \delta-\rho-(\lambda+\kappa)E_{P_n}\!\!\left[f^*\!\!\left(-\frac{\mathcal{L}_\theta+\rho}{\lambda+\kappa}\right)\!\right]\right\}\notag
\end{align}
and in the KL case, we have the further simplification \begin{align}\label{eq:reformulation_final_KL_app}
&\inf_{\theta\in\Theta}\inf_{Q:\mathrm{KL}(Q\|P_n)\leq \delta} \{E_{Q}[\mathcal{L}_\theta]+\kappa \mathrm{KL}(Q\|P_n)\} \\
=&\inf_{\theta\in\Theta}\sup_{\lambda>0}\!\left\{-\lambda\delta-(\lambda+\kappa)\log E_{P_n}[\exp(-\mathcal{L}_\theta/(\lambda+\kappa))]\right\}\,.\notag
    \end{align}
Moreover, the objective function on the right-hand side of \eqref{eq:reformulation_final_app} is concave in $(\lambda,\rho)$ and the objective  on the right-hand side of \eqref{eq:reformulation_final_KL_app} is concave in $\lambda$.
\end{theorem}
\begin{remark}
    Note that when applying the definition \ref{eq:reformulation_final_KL}, the function $f$ should be continuously extended to the endpoints of its domain. 
\end{remark}
\begin{proof}
We will prove  \eqref{eq:reformulation_final_app} by showing that     \begin{align}\label{eq:reformulation_inner}
&\inf_{Q: D_f(Q\|P_n)\leq \delta}\{E_Q[\mathcal{L}_\theta]+\kappa D_f(Q\|P_n)\}\\
=&\sup_{\substack{\lambda>0,\\\rho\in\mathbb{R}} }\!\left\{\!-\lambda \delta-\rho-(\lambda+\kappa)E_{P_n}\!\!\left[f^*\!\!\left(-\frac{\mathcal{L}_\theta+\rho}{\lambda+\kappa}\right)\!\right]\right\}\label{eq:reformulation_inner2}
\end{align}
for each $\theta$.  Once that is established, the KL result \eqref{eq:reformulation_final_KL_app} will then follow by noting that $f^*_{\mathrm{KL}}(y)=e^{y-1}$, which allows \eqref{eq:reformulation_inner2}  to be simplified via the computation
\begin{align}
    &\sup_{\substack{\lambda>0,\\\rho\in\mathbb{R}} }\!\left\{\!-\lambda \delta-\rho-(\lambda+\kappa)E_{P_n}\!\!\left[f_{\mathrm{KL}}^*\!\!\left(-\frac{\mathcal{L}_\theta+\rho}{\lambda+\kappa}\right)\!\right]\right\}\\
    =&\sup_{\lambda>0}\left\{-\lambda \delta-(\lambda+\kappa)\inf_{\rho\in\mathbb{R} }\!\left\{\rho/(\lambda+\kappa)+\exp(-\rho/(\lambda+\kappa)-1)E_{P_n}\!\!\left[\exp\left(-\mathcal{L}_\theta/(\lambda+\kappa)\right)\!\right]\right\}\right\}\notag\\
    =&\sup_{\lambda>0}\!\left\{-\lambda\delta-(\lambda+\kappa)\log E_{P_n}[\,\exp(-\mathcal{L}_\theta/(\lambda+\kappa))\,\,]\right\}\,,\label{eq:app_KL_duality_objective}
\end{align}
where evaluating the supremum over $\rho$ to obtain the last equality is a simple calculus exercise.  Concavity  in $(\lambda,\rho)$ of the objective in \eqref{eq:reformulation_inner2}, i.e., convexity of its negative,  follows from the convexity of the Legendre transform, $f^*$, together with the following properties of convex functions: 1)  the perspective of a convex function is convex, 2) linear functions are convex, 3) a linear combination of convex functions with non-negative coefficients is convex. To see that  the objective in \eqref{eq:app_KL_duality_objective} is  concave in $\lambda$, simply recall that minimizing a jointly convex function over one of its arguments produces  a  function that is convex in the remaining argument.

To complete the proof, we now show the equality \eqref{eq:reformulation_inner}~-~\eqref{eq:reformulation_inner2}.  Let $V$ be the vector space of finite signed measures on $\mathcal{Z}$ and define $W$ to be subset of probability measures satisfying $D_f(Q\|P_n)<\infty$; note that $W$ is a nonempty convex subset of $V$.  Define $F:W\to \mathbb{R}$ by $F[Q]=E_Q[\mathcal{L}_\theta]+\kappa D_f(Q\|P_n)$; note that $D_f(Q\|P_n)<\infty$ implies $Q\ll P_n$, hence  $Q$ is supported on $\{z_i\}_{i=1}^n$, and therefore $\mathcal{L}_\theta\in L^1(Q)$.  $F$ is convex due to the convexity of $f$-divergences;   this follows from the variational representation of $f$-divergences \citep{Broniatowski,Nguyen_Full_2010,birrell2022f}.  The inequality constraint $D_f(Q\|P_n)\leq \delta$ is a convex constraint and is strictly satisfied at $Q=P_n$, i.e., $D_f(P_n\|P_n)=0<\delta$.  Therefore we have shown the the Slater condition holds, see, e.g.,  Theorem 3.11.2  in \cite{ponstein2004approaches} and Theorem 8.3.1 and Problem 8.7 in \cite{luenberger1997optimization}, and hence we have strong duality:
\begin{align}\label{eq:strong_duality_app}
 &\inf_{Q:D_f(Q\|P_n)\leq \delta}\left\{E_Q[\mathcal{L}_\theta]+\kappa D_f(Q\|P_n)\right\}\\
 =&\sup_{\lambda>0 }\left\{-\lambda \delta+\inf_{Q:D_f(Q\|P_n)<\infty}\{E_Q[\mathcal{L}_\theta]+(\lambda+\kappa) D_f(Q\|P_n)\}\right\}\,.\notag
 \end{align}
We emphasize that the ability to restrict the optimization to $\lambda>0$ follows from the fact that the function $Q\mapsto D_f(Q\|P_n)$, which defines the inequality constraint, is bounded below (in fact, it is non-negative).  Finally, for all $\lambda>0$ the inner minimization over $Q$ can be evaluated by using the Gibbs variational formula for $f$-divergences, see Theorem 4.2 in \cite{BenTal2007}, which yields
\begin{align}\label{eq:Gibbs_app}
  &\inf_{Q:D_f(Q\|P_n)<\infty}\{E_Q[\mathcal{L}_\theta]+(\lambda+\kappa) D_f(Q\|P_n)\}\\
  =&-(\lambda+\kappa)\sup_{Q }\{E_Q[-\mathcal{L}_\theta/(\lambda+\kappa)]- D_f(Q\|P_n)\}\notag\\
  =&-(\lambda+\kappa)\inf_{\nu\in\mathbb{R} }\{\nu+E_{P_n}[f^*(-\mathcal{L}_\theta/(\lambda+\kappa)-\nu)]\}\notag\,.
\end{align}
Combining \eqref{eq:strong_duality_app} and \eqref{eq:Gibbs_app}
we arrive at 
\begin{align}
 &\inf_{Q:D_f(Q\|P_n)\leq \delta}\left\{E_Q[\mathcal{L}_\theta]+\kappa D_f(Q\|P_n)\right\}\\
 =&\sup_{\lambda>0 }\left\{-\lambda \delta-(\lambda+\kappa)\inf_{\nu\in\mathbb{R} }\{\nu+E_{P_n}[f^*(-\mathcal{L}_\theta/(\lambda+\kappa)-\nu)]\}\right\}\notag\\
  =&\sup_{\lambda>0,\rho\in\mathbb{R} }\left\{-\lambda \delta-\rho-(\lambda+\kappa)E_{P_n}\left[f^*\left(-\frac{\mathcal{L}_\theta+\rho}{\lambda+\kappa}\right)\right]\right\}\notag\,,
 \end{align}
where to obtain the last line we changed variables from $\nu$ to $\rho=(\lambda+\kappa)\nu$.  Therefore we have shown the claimed equality \eqref{eq:reformulation_inner}~-~\eqref{eq:reformulation_inner2}. This completes the proof.

\end{proof}

\section{Derivation of Eq.~\eqref{eq:delta_rmax} and Interpretation of the $f$-Divergence Neighborhood Size}\label{app:nbhd_interp}

In this appendix we derive the relation  \eqref{eq:delta_rmax} between   $f$-divergence neighborhood size, $\delta$, and the maximum fraction of samples that can be forgotten, denoted $r_{max}$, as previously stated in Section \ref{sec:delta_interp}:
\begin{align}\label{eq:delta_rmax_app}
    \delta=r_{max}f(0) +(1-r_{max}) f(1/(1-r_{max}))\,.
\end{align}
For this derivation, we will assume that  $f(0)<\infty$ and $f$ is $C^1$ and strictly convex (as in the KL and $\alpha$-divergence cases), and that the samples $(x_i,y_i)$ are distinct. 

Suppose that $D_f(Q\|P_n)\leq \delta$. This implies $Q=\sum_i q_i \delta_{(x_i,y_i)}$ for some $q_i\in[0,1]$ that sum to $1$, i.e., $Q$ is  a reweighting of the samples.  By the definition of the $f$-divergence \eqref{eq:f_div_def} we have 
\begin{align}
    D_f(Q\|P_n)=\frac{1}{n}\sum_{i=1}^n f(q_i/n^{-1}) \,.
\end{align}
Consider the case where $Q$ `forgets' $k$ of the samples, with the remainder being equally weighted, i.e., after reordering, $q_i=1/(n-k)$ for $i=1,...,n-k$, $q_i=0$ for $i>n-k$. In such a case we have  
\begin{align}\label{eq:Df_equal_weight}
    D_f(Q\|P_n)=& rf(0) +(1-r) f(1/(1-r))\,,\notag
\end{align}
where $r=k/n$ is the fraction of forgotten samples.  Strict convexity of $f$  implies
\begin{align}
    &\frac{d}{dr}(rf(0) +(1-r) f(1/(1-r)))\\
    =&f(0)-f(z)+zf^\prime(z)>0\,,\notag
\end{align}
where $z=1/(1-r)$. Therefore,  given a choice of  $r_{max}$, if $\delta$ is given by \eqref{eq:delta_rmax_app} then $\{Q:D_f(Q\|P_n)\leq \delta\}$ contains $Q\ll P_n$ that forget up to the fraction $r_{max}$ of samples, but not more.  Note that forgetting $k$ samples and then redistributing the weights unequally between the remaining samples only increases $D_f$ and so this neighborhood will not contain distributions that forget a larger fraction  than $r_{max}$, no matter how the weights are distributed.  This is proven by the following calculation: Let $q_i\in[0,1],i=1,...,n-k$ with $\sum_iq_i=1$.  Then using convexity of $f$ we can compute
\begin{align}
D_f\left(\sum_{i=1}^{n-k} q_i\delta_{x_i}\bigg\|P_n\right)
=&\frac{k}{n}f(0)+\frac{1}{n}\sum_{i=1}^{n-k}f(nq_i)\\
\geq  &\frac{k}{n}f(0)+\frac{n-k}{n}f\left(\frac{n}{n-k}\right)\,.\notag
\end{align}
Comparing the right-hand side with \eqref{eq:Df_equal_weight} we see that unequal weighting can only increase the $f$-divergence, as claimed.

\section{\ant~ Implementation}\label{app:pseudocode}
In this appendix we present pseudocode for our proposed \ant~ method. We use the following notation for the required  objective functions. Fixing $\kappa\geq 0$ and given an array of loss values, $\mathcal{L}_i$, $i=1,...,B$, define
\begin{align}
G_{\mathrm{KL}}(\lambda,\mathcal{L})
\coloneqq&-\lambda\delta-(\lambda+\kappa)\log \left(\frac{1}{B}\sum_{i=1}^B \exp(-\mathcal{L}_i/(\lambda+\kappa))\right)\,,\label{eq:G_KL}\\
G_{f}(\lambda,\rho,\mathcal{L})\
\coloneqq&-\lambda \delta-\rho-(\lambda+\kappa)\frac{1}{B}\sum_{i=1}^Bf^*\!(-(\mathcal{L}_i+\rho)/(\lambda+\kappa))\,.\label{eq:G_f}
\end{align}

\subsection{Pseudocode}
In terms of the objectives \eqref{eq:G_KL} and \eqref{eq:G_f}, the pseudocode for our method is provided in Algorithm \ref{alg:ANTIDOTE}.

\begin{algorithm}
   \caption{ {\bf A}dversarial {\bf N}eural-Network {\bf T}raining with {\bf I}nformation-{\bf D}ivergence-Reweighted {\bf O}bjec{\bf t}iv{\bf e}  (ANTIDOTE)}\label{alg:ANTIDOTE}
   \label{alg:example}
\begin{algorithmic}[1]
   \STATE {\bfseries Input:} Labeled training data $(x_i, y_i)$ (including possible label errors), target model $\phi_\theta$ depending on NN-parameters $\theta$, number of training steps $N$, minibatch size $B$, NN learning rate  $lr_\theta$, loss function $L(\tilde{y},y)$, information divergence $D$ (equal to $\mathrm{KL}$ or $D_f$), neighborhood size $\delta$, penalty strength $\kappa$.
   \STATE {\bf Output:}  Trained model $\phi_\theta$, robust to label noise.
\FOR{$n=1,\ldots, N$}
    \STATE Sample a minibatch $B_n$ of size $B$
    \FOR {$(x_i,y_i) \in B_n$}
        \STATE $\mathcal{L}_{\theta,i}\gets L(\phi_\theta(x_i),y_i)$
    \ENDFOR
    \IF{$D=KL$}
        \STATE $\lambda \gets \text{argmax}_{\lambda\geq0}G_{\mathrm{KL}}(\lambda,\mathcal{L}_\theta)$  
        \STATE $\theta \gets \theta-lr_\theta \nabla_{\theta} G_{\mathrm{KL}}(\lambda,\mathcal{L}_\theta)$
    \ELSIF{$D=D_f$}
        \STATE $(\lambda,\rho) \gets \text{argmax}_{\lambda\geq 0,\rho\in\mathbb{R}}G_{f}(\lambda,\rho,\mathcal{L}_\theta)$  
        \STATE $\theta \gets \theta-lr_\theta \nabla_{\theta} G_{f}(\lambda,\rho,\mathcal{L}_\theta)$
    \ENDIF
\ENDFOR
\end{algorithmic}
\end{algorithm}

While we use a standard SGD method to update the NN parameters $\theta$, the optimization over  $(\lambda,\rho)$  (line 12 in Algorithm \ref{alg:ANTIDOTE}), or simply over  $\lambda$ in the KL case (line 9), deserves special mention. Recalling from Theorem \ref{thm:duality} that the corresponding objective functions $G_f$ \eqref{eq:G_f} and $G_{\mathrm{KL}}$ \eqref{eq:G_KL} are concave in $(\lambda,\rho)$ and $\lambda$ respectively,  one can apply special-purpose (convex) optimization methods to these low-dimensional  problems (2-D and 1-D respectively) to quickly compute accurate solutions at each $\theta$. While it is possible to use alternating gradient steps for $\theta$ and $(\lambda,\rho)$, as is typical in saddle-point (i.e., adversarial) problems,  due to the inexpensive nature of the inner optimization (the values of $\mathcal{L}$ do not change during the optimization over $(\lambda,\rho)$ at fixed $\theta$) and the great importance of having accurate $\lambda$ and $\rho$ values, the small amount extra work required by a convex optimization solver is expected to provide significant performance benefits; we observe this to be the case in practice.  In our numerical experiments we focus our attention on the KL case, where the unique maximizing $\lambda$ is obtained to high accuracy, and with little extra cost, by using the bisection method to solve $\partial_\lambda G_{\mathrm{KL}}=0$ for $\lambda$; see Appendix \ref{app:bisection_KL} for further details.  In the general case, the optimization over $\lambda$ and $\rho$ in \eqref{eq:reformulation_final} can be efficiently computed via a standard convex optimization solver and can be implemented  by   a convex optimization layer \cite{amos2017optnet,agrawal2019differentiable}. In fact, such layers can even be applied to  the primal problem 
\begin{align}
\inf_{Q: D_f(Q\|P_n)\leq \delta} E_{Q}[\mathcal{L}_\theta]\,,
\end{align}
viewed as an optimization 
over the $n-1$-dimensional simplex, though at greater cost than the 1-D or 2-D dual formulations used here.

\subsection{Bisection Method for the  KL Objective}\label{app:bisection_KL}
Here we provide further details on our use of the bisection method to solve $\text{argmax}_{\lambda\geq 0}G_{\mathrm{KL}}(\lambda,\mathcal{L})$ for a given array of loss values $\mathcal{L}$, as required in the KL version of \ant~  (see line 9 in Algorithm \ref{alg:ANTIDOTE}); recall the definition of the objective function $G_{\mathrm{KL}}$ from  \eqref{eq:G_KL}.  We will describe the algorithm for the case where $\kappa>0$; we generally find it advisable to use a (small) positive $\kappa$ as it eliminates numerical issues that can otherwise occur when $\lambda\to 0$.

First note that $G_{\mathrm{KL}}$ is smooth and  concave in $\lambda$ and 
\begin{align}\label{eq:G_KL_limit}
    \lim_{\lambda\to \infty} G_{\mathrm{KL}}(\lambda,\mathcal{L})=-\infty\,.
\end{align}
Therefore   $G_{\mathrm{KL}}$ has a maximizer on $\lambda\geq 0$ and either it occurs at $\lambda=0$ or at a solution to $\partial_\lambda G_{\mathrm{KL}}=0$.  We split the analysis into   the following two cases.

{\bf Case 1:}   $\partial_\lambda|_{\lambda=0} G_{\mathrm{KL}}(\lambda,\mathcal{L})\leq 0$\\
Standard convex analysis theory implies $\partial_\lambda G_{\mathrm{KL}}$ is non-increasing and therefore the in this case we see that $\partial_\lambda G_{\mathrm{KL}}\leq 0$ for all $\lambda\geq0$. Hence $G_{\mathrm{KL}}$ is non-increasing and so its maximum over $\lambda\geq 0$ occurs at $\lambda=0$.  

{\bf Case 2:}  $\partial_\lambda|_{\lambda=0} G_{\mathrm{KL}}(\lambda,\mathcal{L})>0$\\
The limit \eqref{eq:G_KL_limit} implies that $\partial_\lambda G_{\mathrm{KL}}$ becomes negative for $\lambda$ large enough, and hence in this case the derivative changes sign. Therefore a solution to $\partial_{\lambda}G_{\mathrm{KL}}=0$ exists, and any such   solution  is then a maximizer of $G_{\mathrm{KL}}$ over $\lambda\in[0,\infty)$.   In most cases the objective is strictly concave and hence the solution is unique, but the uniqueness is not relevant for our purposes; any solution is guaranteed to be an optimizer due to concavity of the objective.

Our code for solving the optimization in line 9 in Algorithm \ref{alg:ANTIDOTE} follows the above outline. First we check which of the two possible cases holds. In Case 1 we return $\lambda=0$.  In Case 2 we initialize a variable $\lambda_r$ at some positive value and increase it until we observe the derivative changing sign, i.e., until $\partial_\lambda|_{\lambda=\lambda_r} G_{\mathrm{KL}}<0$.  Then we apply a standard bisection method solver (we use \texttt{optimize.bisect} in SciPy) to find a solution to $\partial_\lambda G_{\mathrm{KL}}=0$ in the interval $[0,\lambda_r]$ and  then return that  value of $\lambda$.  For implementation purposes, we require the following formula for the derivative
\begin{align}
   \partial_{\lambda}G_{\mathrm{KL}}(\lambda,\mathcal{L})=-\delta-\log\left(\frac{1}{B}\sum_{i=1}^B e^{-\mathcal{L}_i/(\lambda+\kappa)}\right)-(\lambda+\kappa)^{-1}\frac{\frac{1}{B}\sum_{i=1}^B\mathcal{L}_ie^{-\mathcal{L}_i/(\lambda+\kappa)}}{\frac{1}{B}\sum_{i=1}^Be^{-\mathcal{L}_i/(\lambda+\kappa)}}\,. 
\end{align}

\subsection{Selecting $\delta$ via Penalized Optimization}\label{app:delta_penalized_opt}

In addition to treating $\delta$ as a tunable hyperparameter, we also implement a variant of \ant~ that adaptively adjusts $\delta$ via a penalized optimization procedure. Specifically we modify   \eqref{eq:reformulation_final} to
\begin{align}\label{eq:reformulation_opt_delta}
&\inf_{\theta\in\Theta,r_{max}\in(0,1)}\left\{\inf_{Q: D_f(Q\|P_n)\leq \delta(r_{max})}\{E_Q[\mathcal{L}_\theta]+\kappa D_f(Q\|P_n)\}+V(r_{max})\right\}\\
=&\inf_{\theta\in\Theta,r_{max}\in(0,1)}\sup_{\substack{\lambda>0,\\\rho\in\mathbb{R}} }\!\left\{\!-\lambda \delta(r_{max})-\rho-(\lambda+\kappa)E_{P_n}\!\!\left[f^*\!\!\left(-\frac{\mathcal{L}_\theta+\rho}{\lambda+\kappa}\right)\!\right]+V(r_{max})\right\}\notag
\end{align}
and, similarly, we modify the KL variant \eqref{eq:reformulation_final_KL} to
\begin{align}\label{eq:reformulation_opt_delta_KL}
&\inf_{\theta\in\Theta,r_{max}\in(0,1)}\left\{\inf_{Q:\mathrm{KL}(Q\|P_n)\leq \delta(r_{max})} \{E_{Q}[\mathcal{L}_\theta]+\kappa \mathrm{KL}(Q\|P_n)\} +V(r_{max}) \right\}\\
=&\inf_{\theta\in\Theta,r_{max}\in(0,1)}\sup_{\lambda>0}\!\left\{-\lambda\delta(r_{max})-(\lambda+\kappa)\log E_{P_n}[\exp(-\mathcal{L}_\theta/(\lambda+\kappa))]+V(r_{max})\right\}\,.\notag
    \end{align}

 In \eqref{eq:reformulation_opt_delta} and \eqref{eq:reformulation_opt_delta_KL}, we let $\delta(r_{max})$ be given by \eqref{eq:delta_rmax_app} (for the appropriate $f$) and we choose the penalty $V(r_{max})$ to be a positive increasing function that approaches $\infty$ as $r_{max}\to 1^-$. The introduction of the penalized optimization over $r_{max}$   is based on the intuition that a larger portion of noisy labels will lead to a larger loss, hence the outer minimization in \eqref{eq:reformulation_opt_delta} and \eqref{eq:reformulation_opt_delta_KL} will adaptively  make $r_{max}$ larger to compensate, but it is prevented from sending $r_{max}$ to $1$ by the penalty term.  In practice we work with a penalty of the form
\begin{align}
    V(r_{max})=\frac{Cr_{max}^2}{1-r_{max}}\,,
\end{align}
where $C>0$ is a hyperparameter.  In principle, one could argue we have simply exchanged one hyperparameter ($\delta$) for another ($C$). However, in practice we find that the adaptivity in $\delta$ afforded by the penalized-optimization makes the results much less sensitive to the value of $C$ than \eqref{eq:reformulation_final} and \eqref{eq:reformulation_final_KL} are to $\delta$, and so tuning $C$ is generally much easier than tuning $\delta$. 

In Table \ref{tab:sensitivity_penalty} we provide results on CIFAR-10 using the penalized-optimization variant \eqref{eq:reformulation_opt_delta_KL} of \ant~.  These experiments all used $\kappa = 0.05$. For the asymmetric noise, we used $C=1$ for 20\%, 30\%, 40\% noise ratio and $C=5$ for 10\% noise ratio. For the symmetric case, we used $C=0.05$ for 60\% and 80\% noise ratio; $C=1$ for 20\% noise ratio. These results show that  \eqref{eq:reformulation_opt_delta_KL} also  outperforms \cite{wangepsilon_2024_neurips} in all cases and in multiple cases it also exceeds the  \ant~ variant \eqref{eq:reformulation_final_KL} that uses $\delta$ as a hyperparameter.  As  mentioned  above, and as indicated by the small number of distinct values of $C$ used to obtain these results, the advantage of the penalized-optimization formulation \eqref{eq:reformulation_opt_delta_KL} is that tuning $C$ is generally much easier than tuning $\delta$.

\begin{table*}[h!]
\small
\centering
\setlength\tabcolsep{1 pt}
\caption{Last epoch test accuracies (\%) of the \ant~ variant \eqref{eq:reformulation_opt_delta_KL} on CIFAR-10 at various levels of symmetric and asymmetric noise. All results are reported over 3 random runs (mean±std).}
\label{tab:sensitivity_penalty}
\begin{tabular}{ll|llll|llll}
\toprule
CIFAR-10 &
  \multicolumn{1}{c}{\multirow{2}{*}{Clean}} &
  \multicolumn{4}{c}{Symmetric Noise Rate} &
  \multicolumn{4}{c}{Asymmetric Noise Rate} \\ %%\cline{1-1} \cline{3-10} 
Loss &
  \multicolumn{1}{c}{} &
  \multicolumn{1}{c}{0.2} &
  \multicolumn{1}{c}{0.4} &
  \multicolumn{1}{c}{0.6} &
  \multicolumn{1}{c}{0.8} &
  \multicolumn{1}{c}{0.1} &
  \multicolumn{1}{c}{0.2} &
  \multicolumn{1}{c}{0.3} &
  \multicolumn{1}{c}{0.4} \\ \hline
CE &
  90.50\fontsize{7pt}{7pt}\selectfont±0.35 &
  \multicolumn{1}{l}{75.47\fontsize{7pt}{7pt}\selectfont±0.27} &
  \multicolumn{1}{l}{58.46\fontsize{7pt}{7pt}\selectfont±0.21} &
  \multicolumn{1}{l}{39.16\fontsize{7pt}{7pt}\selectfont±0.50} &
  18.95\fontsize{7pt}{7pt}\selectfont±0.38 &
  \multicolumn{1}{l}{86.98\fontsize{7pt}{7pt}\selectfont±0.31} &
  \multicolumn{1}{l}{83.82\fontsize{7pt}{7pt}\selectfont±0.04} &
  \multicolumn{1}{l}{79.35\fontsize{7pt}{7pt}\selectfont±0.66} &
  75.28\fontsize{7pt}{7pt}\selectfont±0.58 \\
CE${}_\epsilon$+MAE &
  91.40\fontsize{7pt}{7pt}\selectfont±0.12 &
  \multicolumn{1}{l}{89.29\fontsize{7pt}{7pt}\selectfont±0.10} &
  \multicolumn{1}{l}{85.93\fontsize{7pt}{7pt}\selectfont±0.19} &
  \multicolumn{1}{l}{79.52\fontsize{7pt}{7pt}\selectfont±0.14} &
  58.96\fontsize{7pt}{7pt}\selectfont±0.70 &
  \multicolumn{1}{l}{90.30\fontsize{7pt}{7pt}\selectfont±0.11} &
  \multicolumn{1}{l}{88.62\fontsize{7pt}{7pt}\selectfont±0.18} &
  \multicolumn{1}{l}{85.56\fontsize{7pt}{7pt}\selectfont±0.12} &
  78.91±\fontsize{7pt}{7pt}\selectfont0.25 \\ \hline
\textbf{\ant~\eqref{eq:reformulation_final_KL}} &
  {91.91\fontsize{7pt}{7pt}\selectfont±0.06} &
  \multicolumn{1}{l}{\textbf{90.34\fontsize{7pt}{7pt}\selectfont±0.16}} &
  \multicolumn{1}{l}{\textbf{87.71\fontsize{7pt}{7pt}\selectfont±0.17}} & %% cosine: 86.61±0.28
  \multicolumn{1}{l}{\textbf{83.17\fontsize{7pt}{7pt}\selectfont±0.53}} &
  \textbf{60.68\fontsize{7pt}{7pt}\selectfont±1.08}  & %% without bisection: 60.37±0.82
  \multicolumn{1}{l}{\textbf{91.40\fontsize{7pt}{7pt}\selectfont±0.14}} &
  \multicolumn{1}{l}{\textbf{89.90\fontsize{7pt}{7pt}\selectfont±0.06}} &
  \multicolumn{1}{l}{\textbf{87.95\fontsize{7pt}{7pt}\selectfont±0.28}} &
  \textbf{84.60\fontsize{7pt}{7pt}\selectfont±0.50} \\
\textbf{\ant~\eqref{eq:reformulation_opt_delta_KL}} &
  {{92.13\fontsize{7pt}{7pt}\selectfont±0.01}} &
  \multicolumn{1}{l}{\textbf{90.45\fontsize{7pt}{7pt}\selectfont±0.19}} &
  \multicolumn{1}{l}{\textbf{87.63\fontsize{7pt}{7pt}\selectfont±0.29}} & %% cosine: 86.61±0.28
  \multicolumn{1}{l}{\textbf{82.29\fontsize{7pt}{7pt}\selectfont±0.29}} &
  \textbf{\textbf{61.19\fontsize{7pt}{7pt}\selectfont±0.40}}  & %% without bisection: 60.37±0.82
  \multicolumn{1}{l}{\textbf{91.33\fontsize{7pt}{7pt}\selectfont±0.14}} &
  \multicolumn{1}{l}{\textbf{89.66\fontsize{7pt}{7pt}\selectfont±0.28}} &
  \multicolumn{1}{l}{\textbf{87.97\fontsize{7pt}{7pt}\selectfont±0.27}} &
  \textbf{82.97\fontsize{7pt}{7pt}\selectfont±0.39} \\ \hline

\end{tabular}
\end{table*}

\section{Experimental Settings}\label{app:param}

We followed the experimental settings in \citep{wangepsilon_2024_neurips, zhou2021asymmetric_loss, ma2020normalized_normalized}. To ensure reproducibility, we integrated our implementation into the code provided by \cite{wangepsilon_2024_neurips, zhou2021asymmetric_loss}.  We compare the performance of \ant~ with multiple benchmark methods, including standard cross entropy loss (CE), FL \cite{ross2017focal_loss}, MAE loss \cite{ghosh2017robust_MAE_loss}, GCE \cite{zhang2018generalized_GCE_loss}, SCE \cite{wang2019symmetric_SCE_loss}, RCE \citep{wang2019symmetric_RCE_Loss}, APL \cite{ma2020normalized_APL_loss}, AUL \cite{zhou2021asymmetric_loss}, AEL \cite{zhou2021asymmetric_loss}, AGCE \cite{zhou2021asymmetric_loss}, Negative Learning for Noisy Labels (NLNL) \cite{kim2019nlnl_loss}, LDR-KL \cite{zhu2023label_LDR_KL_ICML}, 
LogitClip \cite{wei2023mitigating_LC_loss}, and $\epsilon$-softmax loss \cite{wangepsilon_2024_neurips}. For MAE, RCE, and asymmetric losses (AEL, AGCE, and AUL) we also considered their combination with normalized CE. Following \cite{wangepsilon_2024_neurips,zhou2021asymmetric_loss, ma2020normalized_normalized}, we used an eight-layer convolutional neural net (CNN) for CIFAR-10 and a ResNet-34 for CIFAR-100. The networks were trained for 120 and 200 epochs for CIFAR-10 and CIFAR-100, respectively. The batch size for both CIFAR-10 and CIFAR-100 was $128$. We used the SGD optimizer with momentum 0.9. The weight decay was set to $1e-4$ and $1e-5$ for CIFAR-10 and CIFAR-100, respectively. The initial learning rate was set to $0.05$ for CIFAR-10 and $0.1$ for CIFAR-100. We reduced the learning rate by a factor of 10 at epoch $100$ for both datasets. The gradient norms were clipped to $5.0$ and the minimum allowable value for log was set to $1e-8$. Standard data augmentations were applied on both datasets, including random shift and horizontal flip on CIFAR-10, and random shift, horizontal flip, and random rotation on CIFAR-100. 

For WebVision and ImageNet datasets, the tests follow \cite{wangepsilon_2024_neurips,zhou2021asymmetric_loss, ma2020normalized_normalized}. Consistent with these studies, we used the mini WebVision setting and trained a ResNet-50 using SGD for $250$ epochs with initial learning rate $0.4$, Nesterov
momentum $0.9$ and weight decay $3e-5$ with the batch size of 256. We reduced the learning rate by a factor of 10 at epoch $125$. All images were resized to $224 \times 224$. Typical data augmentations including random width/height shift, color jittering, and horizontal flip were applied. Following \cite{wangepsilon_2024_neurips,zhou2021asymmetric_loss, ma2020normalized_normalized}, we trained the
model on WebVision and evaluated it on the same 50 concepts on the corresponding WebVision and ILSVRC12 validation sets.

%%In line with the benchmark studies, we note that some prior approaches have proposed to mitigate label noise using different techniques such as self-supervised learning \cite{zhou2024l2b,zheltonozhskii2022contrasttodivide_c2d, Li2020DivideMix} based on pseudo-label prediction \cite{sohn2020fixmatch_pseudolabel}, and MixUp \cite{zhang2018mixup}, co-training \cite{yu2019does_COTeachingPlus}. With careful engineering efforts, these techniques can be used in combination to improve the state-of-the-art performance \cite{zhou2024l2b}. 

%%However, it is challenging to attribute the increase in performance to a certain combination of techniques. For a fair comparison, and to avoid unnecessary complications in evaluation, we focus our comparison on approaches that do not use these additional techniques and which do not require clean labels or label correction. Our goal in these experiments is to isolate and compare the effectiveness of leading robust losses in the literature.

To foster reproducibility we provide a detailed list of the (hyper)parameter settings in our experiments below. 

\subsection{Symmetric and Asymmetric Noise on CIFAR-10 and CIFAR-100}
For \ant~ on CIFAR-10, we set $\kappa=0.05$ for all levels of asymmetric noise. We set $\delta=0.1$ for 10\% asymmetric noise, $\delta=0.2$ for 20\% asymmetric noise, $\delta=0.3$ for 30\% asymmetric noise, and $\delta = 0.35$ for 40\% asymmetric noise. We set $\{\kappa=0.07,\delta = 0.02\}$ for clean data, $\{\kappa=0.05,\delta=0.27\}$ for 20\% symmetric noise, $\{\kappa=0.05,\delta=0.57\}$ for 40\% symmetric noise, $\{\kappa=0.05,\delta = 1\}$ for 60\% symmetric noise, $\{\kappa=0.07, \delta = 1.62\}$ for 80\% symmetric noise.

For \ant~ on CIFAR-100, we set $\kappa=0$ for both symmetric an asymmetric noises. We set $\delta=0.2$ for 10\% asymmetric noise, $\delta=0.4$ for 20\% asymmetric noise, $\delta=0.6$ for 30\% asymmetric noise, and $\delta=1$ for 40\% asymmetric noise. We set $\delta=0.04$ for clean data, $\delta=0.27$ for 20\% symmetric noise, $\delta=0.7$ for 40\% symmetric noise, $\delta=1.2$ for 60\% noise, and $\delta=1.7$ for 80\% asymmetric noise.

For benchmark methods, we used the numbers reported in \cite{wangepsilon_2024_neurips} and the same parameter settings, which follows the parameter settings in \citep{ma2020normalized_normalized,zhou2021asymmetric_loss} matching the settings of the original papers for all benchmark methods. For $CE{}_\epsilon+MAE$, we set $\beta = 5$, $m=1e5$, $\alpha = 0.01$ for CIFAR-10 symmetric noise, and $m=1e3$, $\alpha = 0.02$ for asymmetric noise. For CIFAR-100, we set $\beta = 1$, $m=1e4$ and $\alpha \in \{0.1, 0.05, 0.03, 0.0125, 0.0075\}$ for clean, 20\%, 40\%, 60\%, and 80\% symmetric noise, respectively. For asymmetric noise, we used $m = 1e2$, $\alpha \in \{0.015, 0.007, 0.005, 0.004\}$ for 10\%, 20\%, 30\%, and 40\% asymmetric noise, respectively. For $FL{}_\epsilon+MAE$, we set $\gamma=0.1$ and other parameters were set to be the same as those in $CE{}_\epsilon+MAE$. For FL, $\gamma = 0.5$ was used. For GCE, we set $q = 0.7$ for CIFAR-10, and $q = 0.5$ for clean and 20\% symmetric noise ratio, $q=0.7$ for 40\% and 60\% noise ratio, and $q=0.9$ for 80\% asymmetric noise on CIFAR-100. We set $q = 0.7$ for asymmetric noise on CIFAR-100. For SCE, we set $A=-4$, $\alpha = 0.1$, $\beta = 1$ for CIFAR-10, and $\alpha = 6$, $\beta = 0.1$ for CIFAR-100. For APL (MAE, RCE and RCE), we set $\alpha = 1$, $\beta = 1$ for CIFAR-10, and $\alpha = 10$, $\beta = 0.1$ for CIFAR-100. For AUL, we set $a = 6.3$, $q = 1.5$, $\alpha = 1$, $\beta = 4$ for CIFAR-10, and $a = 6$, $q = 3$, $\alpha = 10$, $\beta = 0.015$ for CIFAR-100. For AGCE, we set $a = 6$, $q = 1.5$, $\alpha = 1$, $\beta = 4$ for CIFAR-10, and $a = 1.8$, $q = 3$, $\alpha = 10$, $\beta = 0.1$ for CIFAR-100.
For AEL, we set $a = 5$, $\alpha = 1$, $\beta = 4$ for CIFAR-10, and $a = 1.5$, $\alpha = 10$, $\beta = 0.1$ for CIFAR-100. For LC, we set $\delta = 1$ for clean, 20\%, and 40\%, symmetric noise, $\delta=1.5$ for 60\% and 80\%symmetric noise on CIFAR-10. $\delta = 2.5$ was used for CIFAR-10 asymmetric noise. We set $\delta = 2.5$ for CIFAR-100 asymmetric noise and $\delta = 0.5$ for symmetric noises. For NLNL, the reported results in their original paper was used. For LDR-KL, we set $\lambda = 10$ for CIFAR-10 and 1 for CIFAR-100. 

\subsection{Real-World Noise on WebVision and ImageNet}
For \ant, we set $\kappa=0.05$ and $\delta=0.27$. For benchmark methods, we used the numbers reported in \cite{wangepsilon_2024_neurips} and the same parameter settings, which uses the best settings from the original papers for all benchmark methods. For GCE, we set $q = 0.7$. For SCE, we set $A = -4$, $\alpha = 10$, and $\beta = 1$. For RCE, we set $\alpha = 50$, $\beta = 0.1$. For AGCE, we set $a = 1e-5$, $q = 0.5$. For $NCE+AGCE$, we set $a = 2.5$, $q = 3$, $\alpha = 50$, $\beta = 0.1$. For LDR-KL, we set $\lambda = 1$. For $CE{}_\epsilon+MAE$, we set $m = 1e3$, $\alpha = 0.015$, $\beta = 0.3$.

{
\subsection{Human Annotation-Induced Noise on CIFAR-100N}
We followed the same settings and results reported in \cite{wangepsilon_2024_neurips}. For \ant~, we used the penalized-optimization variant \eqref{eq:reformulation_opt_delta_KL} with $\kappa=3.0$ and $C=1.2$.}

\section{Comparison with Transition Matrix Methods}\label{app:transition_matrix_comp}
As discussed in Section \ref{sec:related_work}, we primarily compare \ant~ with modified loss methods due to their similar algorithm structure and    computational cost.    In Table \ref{tab:T_matrix_comp} below we compare the results from   \ant~ with the FL \cite{patrini2017making}, RW \cite{liu2015classification,xia2019anchor}, and RENT \cite{baedirichlet} methods for utilizing transition-matrix estimates to address label noise. Various approaches can be used to obtain those estimates; we report results using the DualT method \cite{yao2020dual}, though the results using other estimators are similar (see Table 1  in \cite{baedirichlet}).  We find that \ant~ outperforms the transition-matrix-estimator based methods on CIFAR-10 and CIFAR-100. We note that \cite{baedirichlet} uses ResNet34 and ResNet50 for CIFAR-10 and CIFAR-100, respectively, which are more capable than the neural nets used to evaluate \ant~ in accordance with \cite{wangepsilon_2024_neurips} (i.e., a simple CNN for CIFAR-10 and ResNet34 for CIFAR-100).

\begin{table*}[h!]
\small
\centering
\setlength\tabcolsep{1.1 pt}
\caption{Comparison of \ant~ with transition-matrix based methods on CIFAR-10 and CIFAR-100 with symmetric (SN) and asymmetric noise (ASN).   The results for our \ant~ method all used the KL variant \eqref{eq:reformulation_final_KL}.}
\label{tab:T_matrix_comp}
\begin{tabular}{l|l|ll}
\toprule
CIFAR-10 &
   \multicolumn{1}{c}{SN Rate} &
  \multicolumn{2}{c}{ASN Rate} \\ %%\cline{1-1} \cline{3-10} 
Method &
  \multicolumn{1}{c}{0.2} &
  \multicolumn{1}{c}{0.2} &
  \multicolumn{1}{c}{0.4} \\ \hline
  FL  &
  \multicolumn{1}{l}{79.9\fontsize{7pt}{7pt}\selectfont±0.5} &
  \multicolumn{1}{l}{82.9\fontsize{7pt}{7pt}\selectfont±0.2} &
  77.7\fontsize{7pt}{7pt}\selectfont±0.6 \\
RW  &
  \multicolumn{1}{l}{80.6\fontsize{7pt}{7pt}\selectfont±0.6} &
  \multicolumn{1}{l}{82.5\fontsize{7pt}{7pt}\selectfont±0.2} &
  77.9\fontsize{7pt}{7pt}\selectfont±0.4 \\  
RENT  &
  \multicolumn{1}{l}{82.0\fontsize{7pt}{7pt}\selectfont±0.2} &
  \multicolumn{1}{l}{83.3\fontsize{7pt}{7pt}\selectfont±0.1} &
  80.0\fontsize{7pt}{7pt}\selectfont±0.9 \\
  \hline
\textbf{\ant~ }  &
  \multicolumn{1}{l}{\textbf{90.34\fontsize{7pt}{7pt}\selectfont±0.16}} &
  \multicolumn{1}{l}{\textbf{89.90\fontsize{7pt}{7pt}\selectfont±0.06}} &
  \textbf{84.60\fontsize{7pt}{7pt}\selectfont±0.50} \\\hline
\end{tabular}
\begin{tabular}{l|l|ll}
\toprule
CIFAR-100 &
  \multicolumn{1}{c}{SN Rate} &
  \multicolumn{2}{c}{ASN Rate} \\ %%\cline{1-1} \cline{3-10} 
Method  &
  \multicolumn{1}{c}{0.2} &
  \multicolumn{1}{c}{0.2} &
  \multicolumn{1}{c}{0.4} \\ \hline
  FL   &
  \multicolumn{1}{l}{35.2\fontsize{7pt}{7pt}\selectfont±0.4} &
  \multicolumn{1}{l}{38.3\fontsize{7pt}{7pt}\selectfont±0.4} &
  28.4\fontsize{7pt}{7pt}\selectfont±2.6 \\ 
RW   &
  \multicolumn{1}{l}{38.5\fontsize{7pt}{7pt}\selectfont±1.0} &
  \multicolumn{1}{l}{38.5\fontsize{7pt}{7pt}\selectfont±1.6} &
  23.0\fontsize{7pt}{7pt}\selectfont±11.6 \\ 
RENT   &
  \multicolumn{1}{l}{39.8\fontsize{7pt}{7pt}\selectfont±0.9} &
  \multicolumn{1}{l}{39.8\fontsize{7pt}{7pt}\selectfont±0.7} &
  34.0\fontsize{7pt}{7pt}\selectfont±0.4 \\ 
 \hline
\textbf{\ant~} &
  \multicolumn{1}{l}{\textbf{66.95\fontsize{7pt}{7pt}\selectfont±0.47}} &
  \multicolumn{1}{l}{\textbf{65.87\fontsize{7pt}{7pt}\selectfont±0.29}} &
  \textbf{57.89\fontsize{7pt}{7pt}\selectfont±1.14} \\ \hline
\end{tabular}
\end{table*}

{
\section{Parameter Sensitivity and Ablation Analysis of \ant}
\label{app:sensitivity}
There are two parameters the in \ant~ implementation, $\kappa$ and $\delta$ (or $\kappa$ and $C$ in the formulation from Appendix \ref{app:delta_penalized_opt}). A small manual search of the 2-D space induced by these two hyper-parameters yielded the leading performances reported in the paper. Regarding the relationship of $\delta$ and the ratio of label noise, we have provided Equation \ref{eq:delta_rmax} that captures this relationship. We found in our experiments that Equation \ref{eq:delta_rmax} is very useful in practice to provide the initial ballpark of $\delta$. This is another reason why \ant~ enjoys an efficient hyper-parameter tuning.

First, we used our Equation \ref{eq:delta_rmax} to find a proper initial value for $\delta$ (i.e., 0.22) and explored with different $\kappa$ values as shown in the Table \ref{tab:kappa}.
\begin{table}[!h]
\small
\centering
\caption{Hyperparameter sensitivity for $\kappa$ (CIFAR-10,   20\% symmetric noise)}
\begin{tabular}{llllll}
\toprule
$\kappa$    & \multicolumn{1}{c}{0.01} & \multicolumn{1}{c}{0.02} & \multicolumn{1}{c}{\textbf{0.05}} & \multicolumn{1}{c}{0.08} & \multicolumn{1}{c}{0.1} \\ \hline
Accuracy & 88.81                    & 88.88                    & \textbf{89.16}           & 88.80                    & 88.94   \\        \hline
\end{tabular}
\label{tab:kappa}
\end{table}

% Accuracy
% 0.01	88.81%
% 0.02	88.88%
% 0.05	89.16%
% 0.08	88.80%
% 0.1	88.94%

We then fixed $\kappa$ to 0.05 and explored $\delta$
 values around the vicinity of what is suggested by our Equation \ref{eq:delta_rmax} as shown in Table \ref{tab:delta}.
\begin{table}[!h]
\small
\centering
\caption{Hyperparameter sensitivity for $\delta$ (CIFAR-10, 20\% symmetric noise)}
\begin{tabular}{llllllllll}
\toprule
$\delta$      & \multicolumn{1}{c}{0.2} & \multicolumn{1}{c}{0.22} & \multicolumn{1}{c}{0.25} & \multicolumn{1}{c}{\textbf{0.27}} & \multicolumn{1}{c}{0.32} & \multicolumn{1}{c}{0.35} & \multicolumn{1}{c}{0.4} & \multicolumn{1}{c}{0.45} & \multicolumn{1}{c}{0.5} \\ \hline
Accuracy (\%) & 87.87                   & 89.26                    & 89.81                    & \textbf{90.34}                    & 90.16                    & 89.69                    & 89.26                   & 87.70                    & 85.08   \\ \hline               
\end{tabular}
\label{tab:delta}
\end{table}

We note that the results are relatively sensitive to the value of $\kappa$. We also note that the same value for $\delta$ yielded the best result for CIFAR-100 as well for 20\% noise ratio. As expected from our theoretical results, $\delta$
 largely depends on the noise ratio (and is relatively agnostic to other characteristics of the data). Thus, for each noise ratio, $\delta$ can be easily tuned using our Equation \ref{eq:delta_rmax} as a starting point. In addition, the approach of Appendix \ref{app:delta_penalized_opt} further reduces the difficulty of hyperparameter tuning. These results provide clear evidence for efficiency and effectiveness of hyperparameter tuning of \ant.

Finally, we provide an ablation analysis of \ant. Unlike many other counterparts (e.g., $\epsilon$-softmax), we find that the \ant~ loss does not need to be interpolated with CE or MAE losses, rather it provides the best results in isolation. To shed light on this fact, we summarize the results of interpolating our \ant~ loss and traditional CE in Table \ref{tab:ablation}. The ratio 0 amounts to turning off the \ant~ loss and only using CE loss, while the ratio 1.0 denotes turning off the CE loss and using only \ant~ loss.

\begin{table}[!h]
\vspace{-.3cm}
\small
\centering
\setlength\tabcolsep{5 pt}
\caption{Ablation results for interpolating \ant~ and CE on CIFAR-10 with  20\% symmetric noise}
\begin{tabular}{llllllllllll}
\toprule
\begin{tabular}[c]{@{}l@{}} \ant~ Loss\\ Ratio (vs. CE) \end{tabular} & \multicolumn{1}{c}{0}     & \multicolumn{1}{c}{0.1}   & \multicolumn{1}{c}{0.2}   & \multicolumn{1}{c}{0.3}   & \multicolumn{1}{c}{0.4}   & \multicolumn{1}{c}{0.5}   & \multicolumn{1}{c}{0.6}   & \multicolumn{1}{c}{0.7}   & \multicolumn{1}{c}{0.8}   & \multicolumn{1}{c}{0.9}   & \multicolumn{1}{c}{\textbf{1.0}}     \\\hline
Accuracy (\%)                                                    & 75.47 & 76.42 & 76.48 & 77.57 & 77.42 & 77.34 & 77.06 & 77.77 & 78.41 & 81.95 & \textbf{90.34}\\ \hline
\end{tabular}
\label{tab:ablation}
\end{table}

% Ablation Results (Interpolating ANTIDOTE with CE on CIFAR-10 with Noise Ratio of 0.2)

% | Ratio of ANTIDOTE loss | Accuracy |
% | ---------------------- | -------- |
% | 0.0                    | 75.47%   |
% | 0.1                    | 76.42%   |
% | 0.2                    | 76.48%   |
% | 0.3                    | 77.57%   |
% | 0.4                    | 77.42%   |
% | 0.5                    | 77.34%   |
% | 0.6                    | 77.06%   |
% | 0.7                    | 77.77%   |
% | 0.8                    | 78.41%   |
% | 0.9                    | 81.95%   |
% | **1.0**                | **90.34%** |

%         Empirical time comparison of ANTIDOTE and its counterparts (NVIDIA GeForce RTX 4090 with 24 GB memory)

% | Method                         |      Elapsed Time (minutes:second)           |
% |--------------------------------|----------------------------------------------|
% |                                | CIFAR-10 (120 epochs)| CIFAR-100 (200 epochs)|
% | ------------------------------ | -------------------  | --------------------- |
% | CE (non robust)                | 3:36                 | 33:28                 |
% | e_softmax                      | 4:12                 | 34:13                 |
% | ANTIDOTE (Ours)                | 3:52                 | 33:52                 |

}

\section{Additional Figures}
Figure~\ref{fig:losses} shows additional histograms of \ant~ losses for different noise ratios at different stages of training. Recalling from Section \ref{sec:lambda_rho} that, in the $\alpha$-divergence case, samples are forgotten, i.e., do not influence the training, when $-(\mathcal{L}+\rho)<0$, we plot the histograms with the $x$-axis being $-(\mathcal{L}+\rho)$. Figure~\ref{fig:losses}.(a)-(c) shows the  histograms for the 20\% symmetric noise ratio at 50, 100, and 200 epochs. Similarly, Figures~\ref{fig:losses}.(d)-(f) and Figure~\ref{fig:losses}.(g)-(h) show the \ant~ loss histograms for 50\% and 80\% symmetric noise, respectively. The losses corresponding to clean samples are shown in blue and the ones associated with noisy samples are depicted in orange. 
\begin{figure*}[t]
    \centering
    \includegraphics[width=1.0\linewidth]{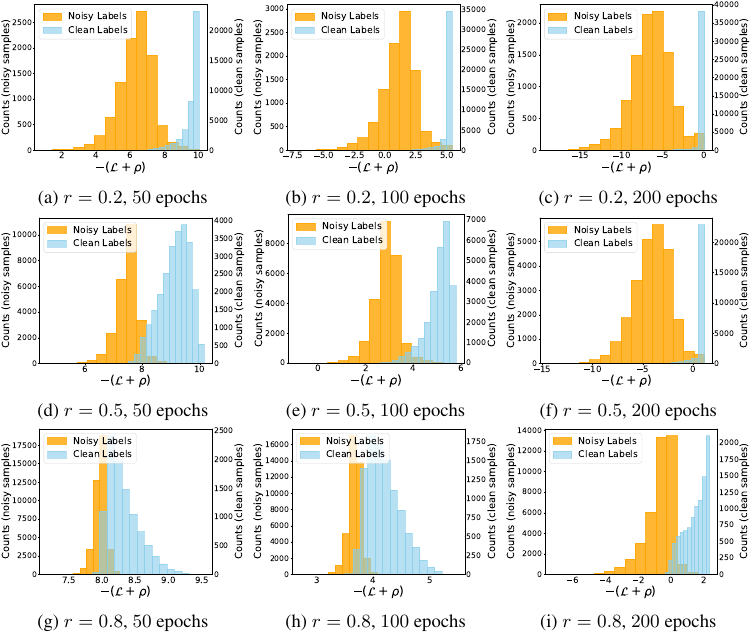}
    \caption{\ant's Loss for Different Noise Ratios During Training}
    \label{fig:losses} \vspace{-4mm}
\end{figure*}

The results here exhibit a similar pattern to the case presented in Section \ref{sec:histogram} above. Specifically,   two key observations are made regarding Figure~\ref{fig:losses}. (1) For all noise levels, the clean loss distribution is to the right side of the noisy loss distribution, implying that \ant~ is able to separate the noisy samples from the clean ones for each of these  noise levels. (2) For each of the noise levels, as the training proceeds from 50 epochs to 200 epochs this separation becomes clearer, with the losses of the samples with noisy labels largely moving below the forgetting threshold at $-(\mathcal{L}+\rho)=0$. These observations collectively suggest that by training with   \ant, our method is effectively learning to distinguish between the clean and noisy labeled samples and is correctly able to determine that the samples with noisy labels should not influence the training of the model.  We again emphasize that the training algorithm was not provided with any (full or partial) information regarding which samples had correct labels.  

%%%%%%%%%%%%%%%%%%%%%%%%%%%%%%%%%%%%%%%%%%%%%%%%%%%%%%%%%%%%

\end{document}